\documentclass[a4paper,fleqn]{cas-sc}

\usepackage[numbers]{natbib}

\usepackage[utf8]{inputenc}
\usepackage{graphicx}
\usepackage{float}
\usepackage{subfig}
\usepackage{url}
\usepackage{colortbl}
\usepackage{amssymb}
\usepackage{pifont}
\usepackage{float}
\usepackage{tablefootnote}
\usepackage{amsmath}
\usepackage[dvipsnames]{xcolor}
\usepackage{placeins}
\usepackage{graphicx}
\usepackage{wrapfig}

\hypersetup{
  colorlinks   = true,
linkcolor=NavyBlue
,citecolor=NavyBlue
,filecolor=Red
,urlcolor=NavyBlue
,menucolor=Red
,runcolor=Red
,linkbordercolor=NavyBlue
,citebordercolor=NavyBlue
,filebordercolor=Red
,urlbordercolor=NavyBlue
,menubordercolor=Red
,runbordercolor=Red
}

\usepackage{algorithm}
\usepackage{algpseudocode}
\usepackage{booktabs}

\newcommand{\revision}[1]{{\color{black}{#1}}}

\def\tsc#1{\csdef{#1}{\textsc{\lowercase{#1}}\xspace}}
\tsc{WGM}
\tsc{QE}

\begin{document}
\let\WriteBookmarks\relax
\def\floatpagepagefraction{1}
\def\textpagefraction{.001}

\shorttitle{Continual Learning in the Presence of Repetition}    

\shortauthors{Hemati et al.\ (2025), to appear in~}  

\title [mode = title]{Continual Learning in the Presence of Repetition}  



%

\author[1]{Hamed Hemati}\corref{cor1}
\ead{hemati.hmd@gmail.com}
\credit{Conceptualization, Software, Investigation, Visualization, Writing – original draft}

\affiliation[1]{organization={Institute for Computer Science, University of St. Gallen},
            addressline={Rosenbergstrasse 30}, 
            city={St. Gallen},
            postcode={9000}, 
            country={Switzerland}
}

\author[2]{Lorenzo Pellegrini}
\credit{Software}
\affiliation[2]{organization={Department of Computer Science, University of Bologna},
            addressline={Via dell’Università 50}, 
            city={Cesena},
            postcode={47521}, 
            country={Italy}}

\author[3,4]{Xiaotian Duan}
\credit{Methodology, Writing – original draft}

\author[3,4]{Zixuan Zhao}
\credit{Methodology, Writing – original draft}

\author[3,4]{Fangfang Xia}
\credit{Methodology, Writing – original draft}
\affiliation[3]{organization={The University of Chicago},
            addressline={5801 S Ellis Ave}, 
            city={Chicago},
            postcode={60637}, 
            country={United States}}

\affiliation[4]{organization={Argonne National Laboratory},
            addressline={9700 S Cass Ave}, 
            city={Lemont},
            postcode={60439}, 
            country={United States}}
            
\author[5,6]{Marc Masana}
\credit{Methodology, Writing – original draft}

\author[5,7]{Benedikt Tscheschner}
\credit{Methodology, Writing – original draft}

\author[5,7]{Eduardo Veas}
\credit{Methodology, Writing – original draft}

\affiliation[5]{organization={Graz University of Technology},
            addressline={Rechbauerstraße 12}, 
            city={Graz},
            postcode={8010}, 
            country={Austria}}
\affiliation[6]{organization={TU Graz - SAL Dependable Embedded Systems Lab, Silicon Austria Labs},
            city={Graz},
            postcode={8010}, 
            country={Austria}}
\affiliation[7]{organization={Know-Center GmbH},
            addressline={Sandgasse 36/4}, 
            city={Graz},
            postcode={8010}, 
            country={Austria}}

\author[8]{Yuxiang Zheng}
\credit{Methodology, Writing – original draft}

\author[8]{Shiji Zhao}
\credit{Methodology, Writing – original draft}

\author[8]{Shao-Yuan Li}
\credit{Methodology, Writing – original draft}

\author[8]{Sheng-Jun Huang}
\credit{Methodology, Writing – original draft}

\affiliation[8]{organization={MIIT Key Laboratory of Pattern Analysis and Machine Intelligence, Nanjing University of Aeronautics and Astronautics},
            city={Nanjing},
            postcode={211106}, 
            country={China}}

\author[9]{Vincenzo Lomonaco}
\credit{Conceptualization}

\affiliation[9]{organization={Department of Computer Science, University of Pisa},
            addressline={Piano Secondo, Largo Bruno Pontecorvo 3}, 
            city={Pisa},
            postcode={56127}, 
            country={Italy}}

\author[10]{Gido M. van~de~Ven}\corref{cor1}
\ead{gido.vandeven@kuleuven.be}
\credit{Supervision, Conceptualization, Visualization, Writing – original draft, Writing – review \& editing}
\affiliation[10]{organization={Department of Electrical Engineering, KU Leuven},
            addressline={Kasteelpark Arenberg 10}, 
            city={Leuven},
            postcode={3001}, 
            country={Belgium}}

\cortext[cor1]{Corresponding authors}



\begin{abstract}
    Continual learning (CL) provides a framework for training models in ever-evolving environments. Although re-occurrence of previously seen objects or tasks is common in real-world problems, the concept of \emph{repetition} in the data stream is not often considered in standard benchmarks for CL. Unlike with the rehearsal mechanism in buffer-based strategies, where sample repetition is controlled by the strategy, repetition in the data stream naturally stems from the environment. 
    This report provides a summary of the CLVision challenge at CVPR 2023, which focused on the topic of repetition in class-incremental learning. The report initially outlines the challenge objective and then describes three solutions proposed by finalist teams that aim to effectively exploit the repetition in the stream to learn continually. The experimental results from the challenge highlight the effectiveness of ensemble-based solutions that employ multiple versions of similar modules, each trained on different but overlapping subsets of classes. This report underscores the transformative potential of taking a different perspective in CL by employing repetition in the data stream to foster innovative strategy design.
\end{abstract}



\begin{keywords}
 Continual learning \sep Class-incremental learning \sep Repetition \sep Competition
\end{keywords}

\maketitle

\section{Introduction}

Designing systems that can learn and accumulate knowledge over time in non-stationary environments is an important objective in artificial intelligence research~\cite{THRUN199525,chen2018lifelong,verwimp2023continual}.
In traditional machine learning, however, the common practice is to build and train models based on statistical learning assumptions. Under these assumptions, a model has access to a static dataset with samples that are independent and identically distributed (IID). This implies that both the training set and test set come from the same distribution and remain unchanged throughout the training and evaluation processes. In reality, these assumptions are often violated, and the model is exposed to different forms of shift in the data. To address such shifts, continual learning (CL) offers a framework to simulate the ``never-ending'' learning setting~\cite{mitchell2018never}. In CL, a model is exposed to a data stream consisting of a potentially unbounded sequence of experiences~\cite{PARISI201954,DELANGE2022,LESORT202052}. Unlike under the statistical learning assumptions, in CL, the model does not have full access to a static dataset. Instead, it gets non-IID access to the data distribution in the form of a data stream, as data become partially available in an incremental manner. In a specific variant of CL, referred to as batch continual learning or task-based continual learning, the model gets \textit{locally IID} access to part of the data distribution. These locally IID parts of the data stream have been referred to as tasks, contexts, or experiences.

A central objective in CL is to design models that mitigate \textit{catastrophic forgetting}~\cite{mccloskey1989catastrophic,ratcliff1990connectionist}. Catastrophic forgetting refers to the phenomenon where a model's performance on a previously learned task rapidly and drastically declines as the model continues to train on data from other tasks. To tackle the issue of forgetting in neural networks, researchers have proposed various strategies. Some strategies add regularization terms to the loss to penalize changes to parts of the model important for past tasks \cite{kirkpatrick2017overcoming, zenke2017continual, li2017learning}. Other strategies replay samples from previous experiences by either employing an external buffer to store raw samples or by using a generative model to synthesize samples similar to those seen previously \cite{rolnick2019experience, chaudhry2019tiny, rebuffi2017icarl, shin2017continual,van2020brain}. While replay strategies often yield good performance, they come with certain drawbacks, such as data privacy concerns and high costs in terms of computation or memory, which can become prohibitive for extended streams. In another CL strategy, parameters are added to increase the capacity of the model to ease the learning of new tasks while avoiding interference with older tasks \cite{rusu2016progressive, aljundi2017expert, xu2018reinforced}.

An aspect that is not often considered in the CL literature is the \textit{repetition} of previously seen concepts over time. CL research often uses ``academic scenarios'' in which tasks, domains or classes are presented in a strictly sequential manner without repetition of previously seen ones~\cite{van2018three,hsu2018re}. (The terms task-, domain- and class-incremental learning are sometimes even interpreted as excluding the possibility of repetition, but see~\cite{van2022three}.) However, in real-world problems, the re-occurrence of previously encountered concepts is common. For example, in a practical scenario where an agent explores a large environment consisting of many areas or rooms over time and learns about new concepts through observation, the agent may revisit similar concepts in different ways as it explores the world. Consequently, the repetition of concepts can naturally happen during the lifetime of the agent, which may lead to an improvement in understanding the concept as it re-appears in different contexts.

Adding the dimension of repetition to a CL problem raises questions such as `\textit{\revision{Does repetition} in a stream affect the behavior of \revision{CL} strategies?}' \revision{and} `\textit{What type of CL strategies are better for streams with repetition?}'.
\revision{It is possible that repetition impacts the performance of different CL strategies in different ways. For example, while replay is typically considered to be a very competitive approach for CL without repetition, it is conceivable that its relative effectiveness might be reduced when there is already naturally occurring repetition in the data stream.}
It is important to point out that there is a fundamental difference between repetition in the data stream and the replay of past experiences from a memory buffer. With replay, the strategy controls the re-occurrence of samples or concepts, and typically, samples from all previous experiences are replayed alongside new data. However, in data streams with repetition, it is the environment or the underlying generation factors of the stream that enforce the recurrence of a concept. Thus, the model might only be presented with a subset of the previously encountered concepts.

\revision{A number of recent works have started to empirically evaluate CL methods in data streams where repetition can happen. To better capture the temporal statistics of real-world data streams, authors have used very long data streams with many tasks in which past classes are re-used~\cite{lesort2022scaling}, data streams with blurry and stochastic task boundaries~\cite{koh2022online,moon2023online}, and data streams modeled based on how infants interact with toys~\cite{stojanov2019incremental}.}
The role and importance of repetition for learning have also been investigated from a neuroscience perspective. For instance, reference~\cite{zhan2018effects} explores how repetition can affect memory performance and brain activity for associative memory over time. Other studies focus on the spacing effect~\cite{smith2017spacing, feng2019spaced}, which concerns how spaced repetition can enhance memory retention compared to massed repetition\revision{, or on the Hebb repetition effect~\cite{cho2024neuromimetic}, which involves the enhancement of memory after each repetition}.

In recent years, several workshops on CL have organized challenges. For example, the first edition of the CLVision challenge at CVPR~2020 introduced new benchmarks based on the CORe50 dataset~\cite{lomonaco2022cvpr}. Although the `New Instances and Classes' scenario \cite{maltoni2019continuous} that was used in that challenge includes some form of repetition of past classes, it lacks a principled approach to the study of the problem. In the 2021 edition of the CLVision challenge, the focus was mainly on supervised class-incremental classification and continual reinforcement learning~\cite{challenge21}. In that same year, in a challenge at ICCV~2021, there was a track on domain-incremental object classification and detection based on a benchmark for autonomous driving called CLAD~\cite{verwimp2022clad}. Finally, the CLVision challenge at CVPR~2022 \cite{pellegrini20223rd} emphasized the design of strategies for object detection and classification using the large EgoObjects dataset~\cite{zhu2023egoobjects}.

For the CLVision challenge at CVPR~2023, the primary goal was to explore the role of repetition in CL, a topic not often explored by the community. To do so, the participants were asked to design strategies that could exploit the repetition inherent in the stream to boost knowledge transfer and reduce forgetting without storing raw samples. This report provides a summary of this challenge. Details about the design of the challenge are provided in section~\ref{sec:challenge}. Solutions proposed by finalist teams are presented in sections~\ref{sec:xduan} to~\ref{sec:pddbend}. Section~\ref{sec:results} presents the results of the challenge, and section~\ref{sec:discussion} discusses the main lessons learned.

\section{Challenge Details}
\label{sec:challenge}

Participants in the challenge were asked to design strategies for a class of CL problems referred to as class-incremental with repetition (CIR). CIR encompasses a variety of streams with two key characteristics:
\begin{enumerate}
\item New classes can appear over time.
\item Previously encountered classes can re-occur with varying repetition patterns.
\end{enumerate}
Since many existing strategies in the CL literature have only been tested in repetition-free settings, their efficacy and performance in the presence of repetition remain unclear. To investigate the influence of repetition in the data stream on the relative effectiveness of different strategies, a set of CIR benchmarks was generated using a sampling-based data stream generator controlled by four interpretable parameters (see subsection~\ref{sec:stream_generator}). Participants were tasked with developing strategies that, upon completing training on the entire stream, could achieve high average accuracy on a test set containing unseen samples from the same classes present in the stream. To allow participants flexibility in strategy design without the need for significant computational resources, we used CIFAR-100~\cite{krizhevsky2009learning}, a relatively small-scale dataset of natural images ($32\times32$ RGB images, 100~classes), as the base dataset for the stream generator.

The competition consisted of two phases:

\paragraph{Pre-selection Phase} The pre-selection phase took place between the 20th of March 2023 and the 20th of May 2023. For this phase, there were three challenge data streams, which were released on the GitHub repository\footnote{\href{https://github.com/ContinualAI/clvision-challenge-2023}{https://github.com/ContinualAI/clvision-challenge-2023}} of the challenge. The participants were asked to submit their solutions in the form of model predictions on the test set after training on each stream. Appendix~\ref{sec:participation_over_time} provides details on the participation over time in the pre-selection phase. 

\paragraph{Final Evaluation Phase} The final phase started from the 20th of May 2023. In this phase, the five highest-ranking participants from the pre-selection phase were asked to send the source code of their strategies. Their strategies were tested on three new data streams, and the average accuracy on the test set after training on each of these three new streams was used as the metric for the final ranking. Participants were allowed to make changes to their strategies between the pre-selection phase and the final evaluation phase.

\subsection{Stream Generator}
\label{sec:stream_generator}

The sampling-based generator developed by Hemati et al.~\cite{hemati2023class} was used to generate the data streams for this challenge using four control parameters with clear interpretations:

\begin{itemize}
    \item Stream length ($N$): Number of experiences in the stream.
    \item Experience size ($S$): Number of samples in each experience. 
    \item First occurrence distribution ($P_f$): A discrete probability distribution over the experiences in the stream that determines the first occurrence of each class.
    \item Repetition probability ($P_r$): Per-class repetition probabilities that control the likelihood of each class re-appearing in future experiences in the stream after its first occurrence. 
\end{itemize}

\begin{wrapfigure}{r}{0.5\textwidth}
\vspace{-0.45cm}
\centering
\includegraphics[width=0.48\textwidth]{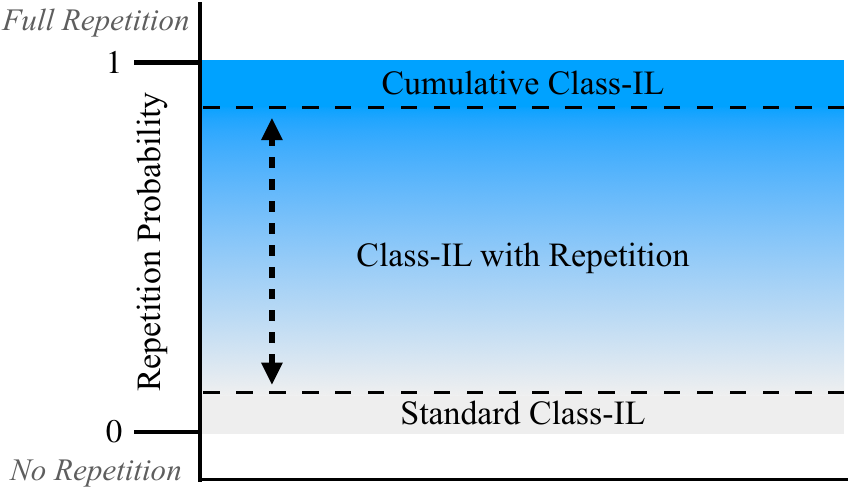}
\caption{Illustration of transitioning from ``standard Class-IL'' to ``cumulative Class-IL'' by increasing the probability of repetition for seen classes, which is set by control parameter~$P_r$. Class-IL: class-incremental learning.}
\vspace{-0.65cm}
\label{fig:cir_repetition_level}
\end{wrapfigure}

By default, an equal number of samples is assigned to all classes present in each experience. $P_f$, which controls the ``first occurrence" property of the stream, determines when dataset classes make their initial appearance. For instance, in one stream, all classes might appear for the first time at the outset, whereas in another stream, new classes might appear gradually. Once a class has appeared for the first time, it re-appears based on a per-class repetition probability, which is controlled by $P_r$. Figure~\ref{fig:cir_repetition_level} illustrates how increasing the repetition probability from $0$ to $1$ corresponds to a transition from ``standard'' class-incremental learning, where each class is present in only one experience, to ``cumulative'' class-incremental learning, where each class re-appears in each experience after its first occurrence. While the repetition probability can be dynamically set based on a time-varying probability mass function, in this challenge, the repetition probability for each class remains fixed over time. This means that, in this challenge, $P_r$ is a list of probability values, with one value assigned to each class in the dataset.

\begin{figure}[!htb]
    \centering
    \subfloat[$P_f: \text{Geom}(0.01)$, $P_r: \text{Zipf}(0.5)$]{
        \includegraphics[width=0.32\textwidth]{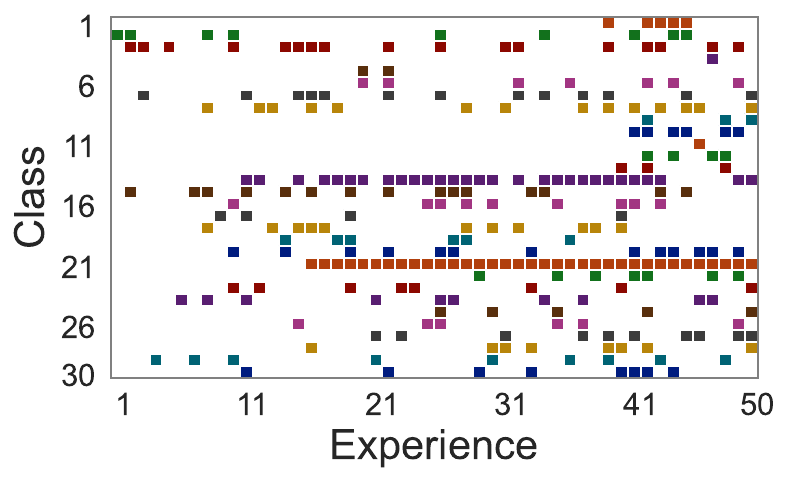}
        \label{fig:subfig1}
    }
    \hfill
    \subfloat[$P_f: \text{Geom}(0.3)$, $P_r: \text{Zipf}(0.5)$]{
        \includegraphics[width=0.32\textwidth]{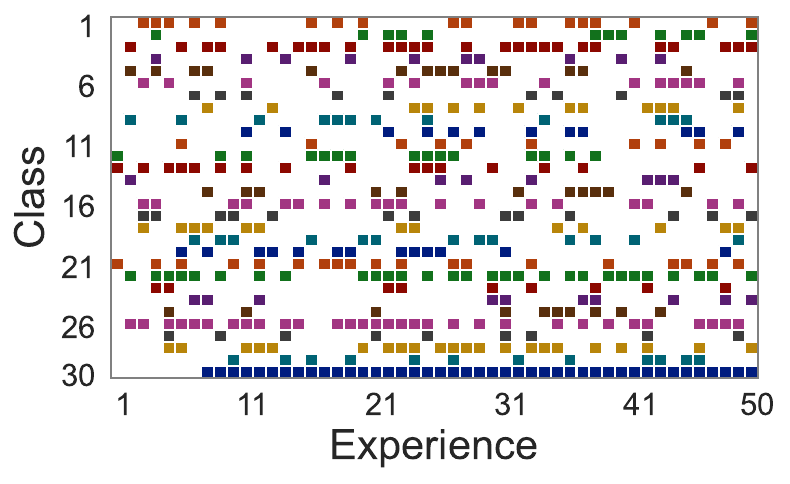}
        \label{fig:subfig2}
    }
    \hfill
    \subfloat[$P_f: \text{Geom}(0.9)$, $P_r: \text{Zipf}(0.5)$]{
        \includegraphics[width=0.32\textwidth]{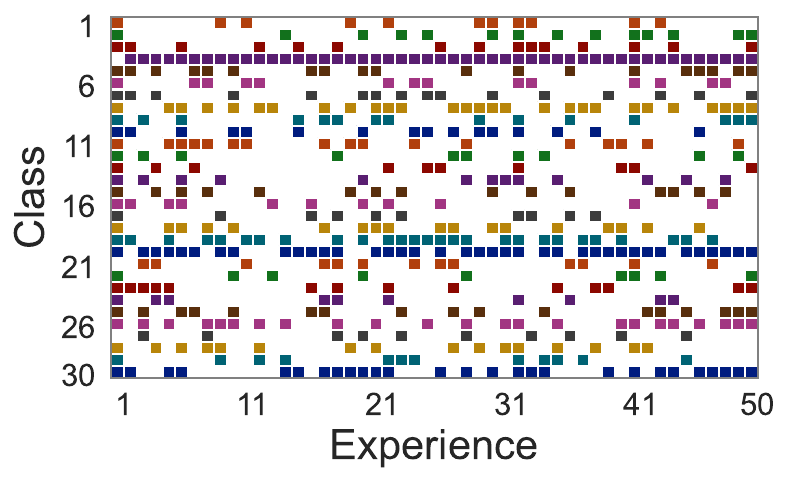}
        \label{fig:subfig3}
    }
    \caption{Examples of generated streams with a Geometric distribution over the first occurrences of the dataset's classes, and a Zipfian distribution for the repetition of classes along the stream.}
    \label{fig:sample_streams}
\end{figure}

In Figure \ref{fig:sample_streams}, three different examples of streams generated using the generator are shown, with a dataset containing 30~classes and $N$ set to $50$. In these examples, $P_f$ is varied while $P_r$ is kept constant to illustrate the effect of first occurrence on the generated streams.

\paragraph{Selected Challenge Stream Parameters}
The settings used for $P_f$ and $P_r$ in each of the challenge data streams are provided in Appendix~\ref{sec:stream_configurations}. The number of experiences per stream ($N$) was fixed as $50$, and the number of samples per experience ($S$) was set to $2000$. Following the default setting, the number of samples in each experience was equally distributed across the classes present in the experience.

\subsection{Participation}
Participation was in the form of teams. Team registrations and solution submissions were handled using the CodaLab platform \cite{JMLR:v24:21-1436}. Each team was allowed to create only one account on the platform to prevent parallel submissions. Participants were provided with a DevKit based on the Avalanche library \cite{carta2023avalanche}: \href{https://github.com/ContinualAI/clvision-challenge-2023/}{https://github.com/ContinualAI/clvision-challenge-2023/}. The DevKit contains all necessary scripts and codes to load the benchmarks for the challenge configurations. Researchers who are interested in evaluating the performance of their own strategies can use the DevKit for both pre-selection and final phase streams.

\subsection{Rules and Restrictions}
\label{sec:rules}

\textbf{Submission:} Each submission in the pre-selection phase consisted of three sets of predictions. Each set of predictions was made by testing the model against the challenge's test set, after training the designed strategy on one of the three challenge streams.

\textbf{Strategy Design:} Within each experience in a data stream, users had full access to the training data from that experience, but not to the data from other experiences. In the default settings of the DevKit, the number of epochs per experience was set to 20. Participants had the flexibility to tweak and tailor the epoch iterations and dataset loading. For instance, a possible strategy would be to iterate for more epochs in the initial experiences and fewer in the final ones.

\textbf{Model Architecture:} All participants were required to utilize the (Slim-)ResNet-18 provided in the DevKit as the base architecture for their models. However, they were allowed to use multiple copies and to add additional modules, e.g. gating modules, provided they adhered to the maximum GPU memory usage (4000 MB) allowed for the competition during training. The GPU memory limit is set to almost eight times the memory size needed to train the Slim-ResNet-18 backbone with a batch size of 64. This allows for considerable flexibility in adding additional modules to the backbone.

\textbf{Replay Buffer:} Using a replay buffer to store dataset samples was not allowed. However, buffers could be used to store any form of data representation, such as the model's internal representations of data samples. Irrespective of the buffer type, the buffer size (i.e., the total number of stored exemplars) was not allowed to exceed 200.

\textbf{Hardware Limitations:} To encourage a comparison between strategies that use similar computational resources, participants were limited to using one GPU for training, and the maximum training time for each run was capped at 500 minutes. No computing limitations were specified for the inference time.

\section{Strategy 1: HAT-CIR}
\label{sec:xduan}

The strategy proposed by team \emph{xduan7} is called \textbf{HAT-CIR}. This strategy combines the strengths of network replicas and test-time decision-making, along with other elements, such as Hard Attention to the Task~(HAT)~\cite{serrà2018hat} and Supervised Contrastive Learning~(SupCon)~\cite{khosla2020supcon}.

\subsection{Motivation and Related Work}

In CIR, as more generally with class-incremental learning, comparing model output logits across different classes is difficult because images from different classes are not jointly used for training. One potential solution is using a memory buffer and replaying samples to calibrate the logits, but due to restrictions on the storage of data, this approach is not well suited for this challenge.
Another strategy involves parameter isolation. Parameter-isolation methods divide the neural network into segments, each responsible for a specific task or experience, preventing the network from forgetting knowledge acquired from prior experience. For example, HAT employs trainable binary masks to partition the network based on the experience identifier. This has proven effective, particularly in task-incremental learning problems where the experience identifier is provided during both training and test phases. However, the parameter-isolation strategy encounters challenges when applied to the CIR setting, where experience identifiers are unavailable during testing.

To address the experience identifier limitation in parameter-isolation methods within the CIR setting, OOD detection techniques can be used to enable ``task-agnostic inference''. Such techniques can detect samples that are not from the current experience, as for example demonstrated in~\cite{kim2022clom} and~\cite{kim2022more}. Another option is to use representation learning methods like SupCon to make the output logits more comparable across different classes. This has been explored in previous works (e.g.,~\cite{kim2022clom}) and has been shown to be effective when applied to the class-incremental learning setting.

In light of these existing works, and considering the specific challenges posed by the competition, the subsequent sections will elaborate on how to adapt and integrate these strategies to propose a novel solution for the CIR setting.

\subsection{Method Description}

\begin{figure}[t]
\centering
\includegraphics[width=0.99\textwidth]{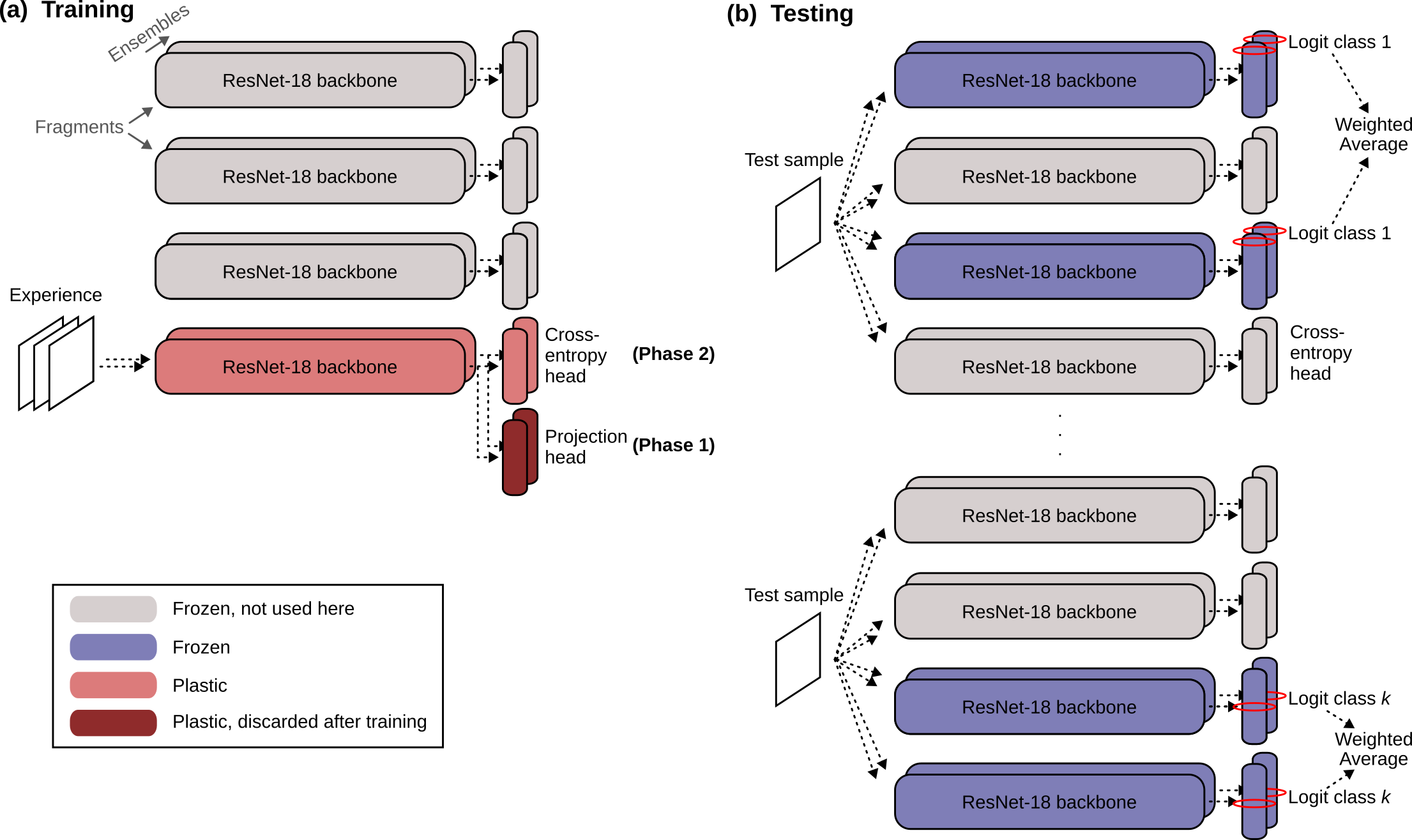}
\vspace{0.2cm}
\caption{\revision{Schematic of HAT-CIR during training and test time. HAT-CIR is illustrated here with network replicas and without HAT-based partitioning. \textbf{(a)} When training on a new experience, a new `fragment' -- consisting of multiple `ensembles' -- is added to the model and trained on the training data of the new experience in two phases. In the first phase, a projection head is used and a supervised contrastive loss is optimized; in the second phase, a softmax output layer is used and a cross-entropy loss is optimized. \textbf{(b)} During testing, a score for each possible class is computed as a weighted average of the logits from the most recent fragments that were trained on an experience in which that class appeared.}}
\vspace{-0.2cm}
\label{fig:hat-cir}
\end{figure}

The proposed method comprises three core components: (i) structural design, featuring HAT-based partitioning and network replicas; (ii) a two-phase training strategy, including supervised contrastive learning and classification; and (iii) a momentum-based inference mechanism for test-time decision making. \revision{A schematic of the method is shown in Figure~\ref{fig:hat-cir}.}

\subsubsection{Structural Design}

\paragraph{HAT-based Partitioning}

To mitigate catastrophic forgetting, HAT isolates network parameters based on experience IDs. However, the original HAT method suffers from  slow training and hyperparameter sensitivity when handling a large number of experiences. To overcome this, HAT-CL~\cite{duan2023hatcl} is used, which initializes masks to ones and uses a cosine mask scaling curve to facilitate better alignment with network weights. By using the cosine mask scaling curve, each training epoch is divided into three phases: 
\begin{enumerate}
    \item  Train weights while masks are mostly ones
    \item  Train masks and weights together to make masks more sparse
    \item  Fine-tune weights while masks are mostly ones again
\end{enumerate}
These changes significantly improve the training speed and stability, and performance of HAT. Furthermore, the effect of the regularization term for the masks is gradually reduced. This step ensures full network capacity utilization and provides an accounting for the number of classes in each experience through variable regularization terms. It is important to note that the HAT-based partitioning was only used in the pre-selection phase; for the final phase, only network replicas were used, which led to higher performance.

\paragraph{Network Replicas}
Two types of network replicas, \textit{fragments} and \textit{ensembles}, are used to increase the model capacity and improve performance. Fragments refer to network replicas trained on different experiences.
For example, if there are ten experiences, five fragments might be trained, each on two experiences.
Ensembles are replicas trained on the same experiences.
For each experience, multiple ensemble replicas will be trained on the same samples with different initializations and data augmentation. Thus, their predictions are averaged for final decision-making. These two replica types are complementary and can be combined with or without HAT-based partitioning. Importantly, adding fragment replicas does not increase the computational costs during training, while adding ensemble replicas does. Due to the challenge's restriction on memory usage and training time, the final version of the model used for the experiments employed 50~fragments and two ensembles. It is important to note that during the pre-selection phase, network replicas were not yet used.

\subsubsection{Two-phase Training}
To learn better feature representations, in each experience the network is first trained with supervised contrastive learning. The objective is to maximize the similarity between feature vectors of the same class and minimize those of different classes with the following loss function:
\[
L = \sum_{i=1}^{N} \max \left( 0, D(f(x_i^a), f(x_i^p)) - D(f(x_i^a), f(x_i^n)) + \alpha \right)
\]
In this equation, $f(x)$ produces the embedded feature vector for input $x$, $D(x,y)$ denotes a distance function, $x_i^a$, $x_i^p$, and $x_i^n$ represent the anchor, positive, and negative samples, respectively, and $\alpha$ is a margin parameter.
The number of samples in a batch is denoted by $N$. The training of hard attention masks occurs exclusively in this supervised contrastive learning phase, due to their sensitivity to learning rates and epoch numbers.

In the second training phase of each experience, the parameters of the network are trained further using a standard cross-entropy classification loss.

\subsubsection{Momentum-based Test-time Decision Making}
\revision{At test time, a predicted ``score'' for each possible class is calculated as a specific weighted average of the logits of that class from different model ``snapshots'', whereby a model snapshot is a fragment or a HAT-based partitioning. In particular, an average is taken over the logits of the $N$ most recent model snapshots that were trained on an experience in which that class appeared.}
This proposed approach, called ``momentum-based decision making'', achieves strong performance in this challenge by setting $N=3$, using the weights $\{1,2,3\}$ for the last three experiences in which each class appeared.

\subsubsection{Other Details}
Standard techniques such as data augmentation, early stopping, and learning rate scheduling were explored. Notably, it was found that data augmentation during both training and inference was crucial for improving performance. For the final submission, a combination of \texttt{RandomResizedCrop}, \texttt{RandomHorizontalFlip}, and \texttt{ColorJitter} transformations was used.
Further details, consisting of an ablation study and an analysis of the influence of the number of network replicas, can be found in Appendix~\ref{sec:appendix_xudan}.
The implementation of \mbox{HAT-CIR} publicly available on GitHub: \href{https://github.com/xduan7/clvis-chlg-2023/}{https://github.com/xduan7/clvis-chlg-2023/}.

\subsection{Discussion}

In this section, the limitations and possible improvements for the method are critically assessed, as well as the implications of the findings for future research.

\paragraph{Limitations}

A significant drawback of the approach arises when the number of classes is small in the initial experiences.
In such cases, the network fails to learn effective class representations due to the limited diversity.
This negatively impacts the performance of the momentum-based test-time decision-making strategy. Another inherent limitation is related to the use of HAT. The rigidity of the network structure requires careful hyperparameter tuning to match the expected total number of experiences, which is not always known in advance in CIR settings. As a result, suboptimal parameter allocation might be achieved, leading to compromised performance.

\paragraph{Future Directions}

The experiments show that the method's performance scales positively with the number of network replicas, implying that network capacity plays a significant role.  Utilizing larger networks, possibly incorporating different structures with pre-trained weights, might offer avenues for further improvement in performance.

\section{Strategy 2: Horde}
\label{sec:mmasana}

The strategy proposed by team \emph{mmasana} is called \textbf{Horde}. This strategy learns an ensemble of feature extractors (FEs) on selected experiences, which should provide robust features useful for discriminating between seen and unseen downstream classes. After training, the learned FEs are frozen to obtain a zero-forgetting ensemble. However, since not all classes are present in each experience, the outputs need alignment to balance them effectively. To do this, Horde uses pseudo-feature projection on the outputs to retain as much discrimination between classes as possible, while learning a unified head capable of discriminating between all classes seen so far. To further facilitate the pseudo-feature projection, FEs are trained with both the usual cross-entropy loss and an additional metric learning loss, which promotes alignment among the learned classes within each feature space. 

\subsection{Motivation and Related Work}
An important motivation for the proposed approach is balancing the stability-plasticity dilemma~\cite{mermillod2013stability}. This balance involves retaining useful knowledge from previous experiences while learning new ones from the sequence.
For plasticity, robust and discriminative representations~\cite{oquab2014learning} are important, as they help adapt to new classes and promote generalization (or forward transfer). Horde encourages the learning of such representations by combining a metric learning loss and the usual cross-entropy classification loss.
Support for stability can be provided by zero-forgetting methods~\cite{masana2022class}, which freeze the feature extractor part of model after training~\cite{aljundi2017expert, rusu2016progressive, rajasegaran2019random} or apply masks to the parameters or the intermediate representations~\cite{fernando2017pathnet, mallya2018packnet, masana2020ternary}. The set up of the challenge excludes the use of the experience ID at test time and the use of pretrained models, both of which limit the direct application of existing zero-forgetting methods. However, the enforcement of the stability component via freezing is preserved for Horde and represented with the use of an ensemble of FEs that retain all learned knowledge.

Initial inspiration for Horde is provided by the existing methods Feature Translation for Exemplar-Free Class-Incremental Learning~(FeTrIL)~\cite{petit2023fetril} and Ternary Feature Masks~(TFM)~\cite{masana2020ternary}. FeTrIL~\cite{petit2023fetril} uses a pretrained frozen backbone to benefit from the stability of having zero-forgetting for that part of the model, while the plasticity is introduced via the learning of a unified head. This method has a pseudo-feature generator that uses the representations from a single FE and applies a geometric translation to align both past and new classes. The pretrained backbone could be replaced by learning on the first experience, although, the method heavily relies on the initial training covering a majority of classes such that the resulting feature extractor is expressive and discriminative enough. TFM~\cite{masana2020ternary} freezes the existing model and extends it at each experience while reusing the representations from previous experiences. This method applies masking on the outputs of each layer to control the flow of gradients so that previous knowledge can be used but its corresponding weights are not modified. However, the dynamic architecture requires the experience ID at test time. Both methods break the challenge rules in multiple ways when used as proposed in their original works.

Horde is inspired by these two methods by having a dynamic zero-forgetting architecture for the ensemble and making use of a pseudo-feature generation approach for learning the unified head. The approach is adapted to the challenge by extending the pseudo-feature generation with the usage of the standard deviation, and using the contrastive loss~\cite{kulis2013metric, Hadsell06} to improve the learned shape of the representations, and avoiding the issues for the pseudo-feature generation. Furthermore, the proposed pseudo-feature generation method is adapted to the challenge scenarios that contain class repetition.

\begin{figure}[t]
    \centering

    \includegraphics[trim= 0 35 0 35, clip, width=0.47\textwidth]{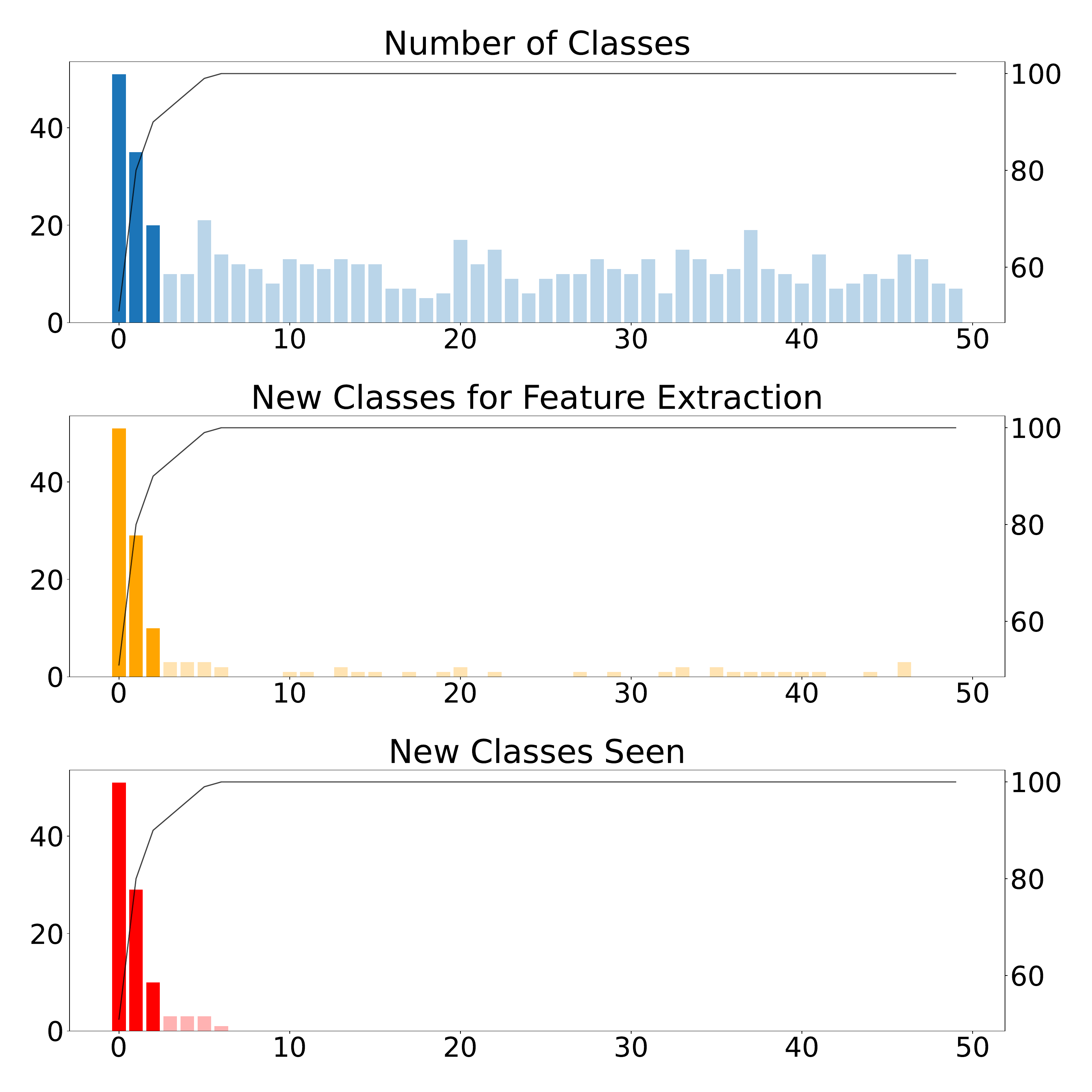}\hfill
    \includegraphics[trim= 0 35 0 35, clip, width=0.47\textwidth]{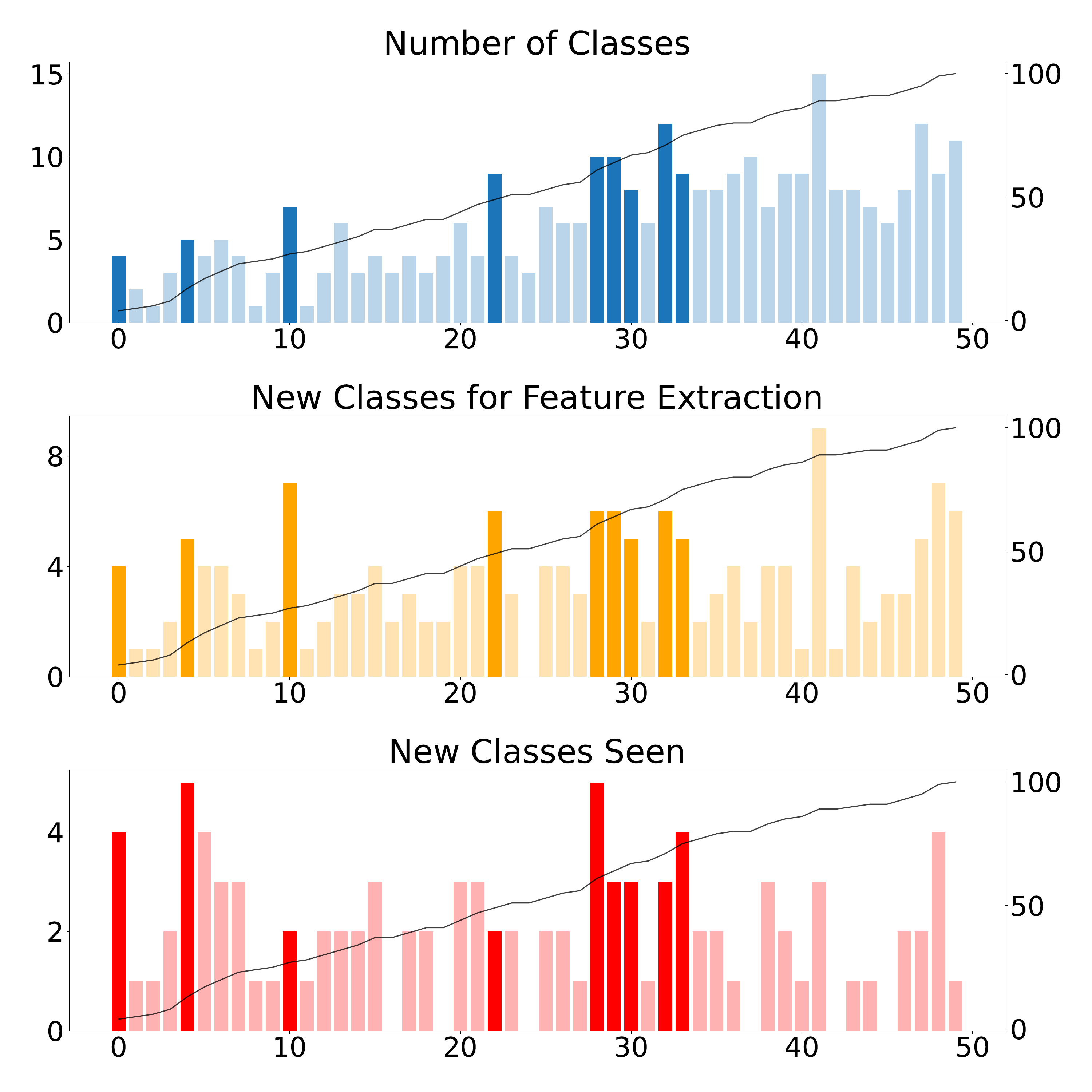}    
    \vspace{0.2cm}
    \caption{The panels indicate for stream 1 (left) and stream 2 (right) in which experiences Horde decided to add a new FE (indicated by bright colors), while comparing the experiences based on three factors: the total number of classes in the experience (blue), the number of classes in the experience on which no FE had been trained yet (yellow), and the number of new, unseen classes in the experience (red). The black line represents the total amount of classes seen so far.}
    \label{fig:horde-heuristic}
    \vspace{-0.3cm}
\end{figure}

\subsection{Method Description}

The proposed strategy Horde combines the feature representations of individual FEs into a single unified head capable of predicting all classes seen so far. This is achieved through a two-step training process: first, the learning of an FE (on selected experiences only), and second, the pseudo-feature alignment for adapting the unified head (on each experience). Each individual FE is an expert model trained on a single experience, after which it is frozen and added to the ensemble. In the second training step, data are passed through all the ensemble's models, and the unified head is fine-tuned. Training the unified head involves using representations directly from the expert FEs familiar with a class and the pseudo-feature projections from the representations of FEs not trained on that class.

\begin{figure}[t]
\centering
\includegraphics[width=0.99\textwidth]{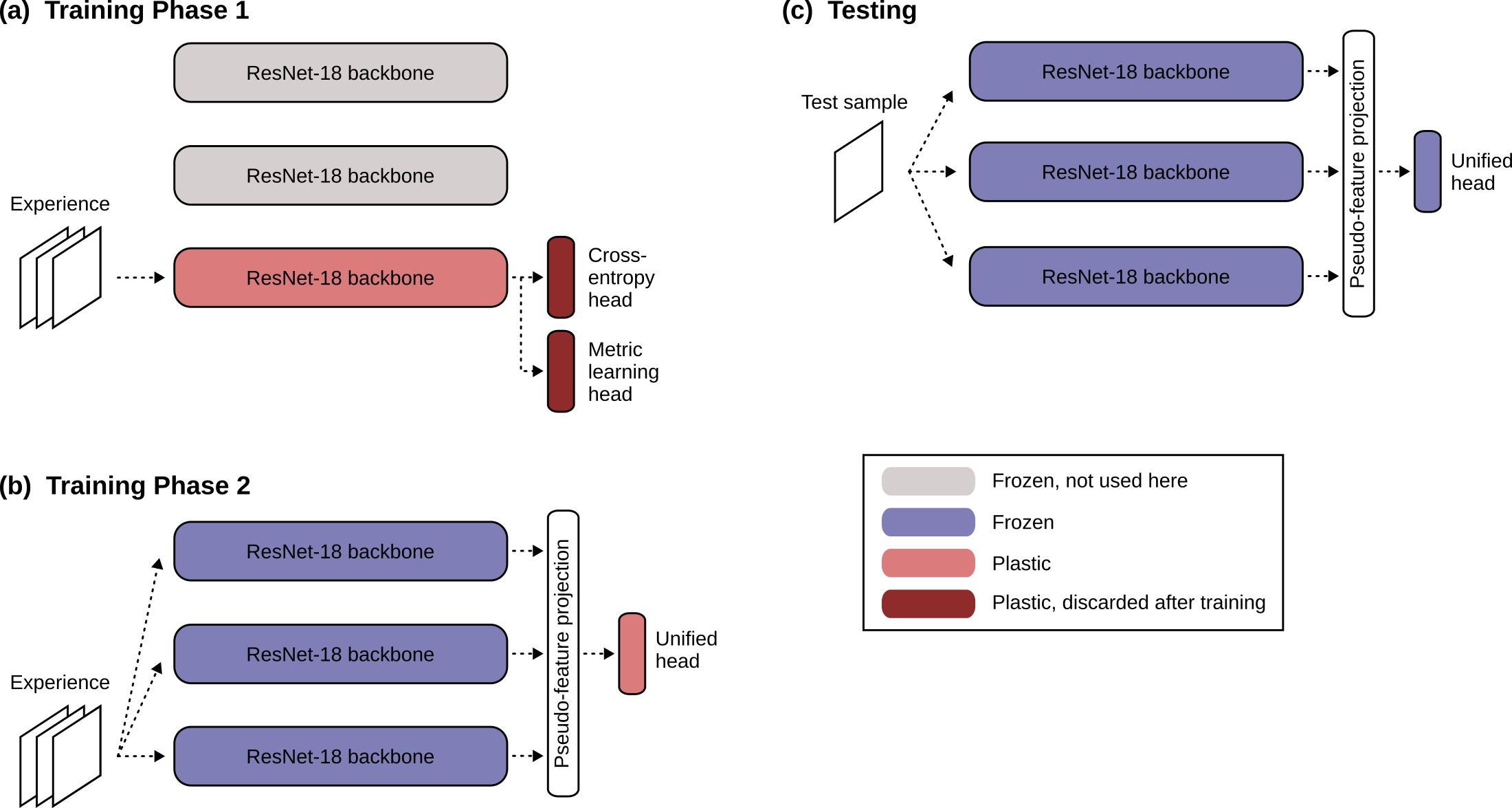}
\vspace{0.2cm}
\caption{\revision{Schematic of Horde during training and test time. \textbf{(a)} When training on a new experience, in the first phase, a new feature extractor might be trained using both a cross-entropy and a contrastive loss. Whether a new feature extractor is trained on the new experience is decided by a heuristic, see Figure~\ref{fig:horde-heuristic}. \textbf{(b)} In the second training phase, which is performed on each new experience, a pseudo-feature projection is performed and a unified head is trained to discriminate between all seen classes based on the features from the ensemble. \textbf{(c)} At test time, a test sample is simply forwarded through all components of the model, and the predicted class is read out from the unified head.}}
\vspace{-0.2cm}
\label{fig:horde}
\end{figure}

\textbf{When to add a new FE to the ensemble:} For the specific settings of the competition, two constraints (heuristics) were designed to determine when to add an FE to the ensemble. First, to constrain the presence of overly overfitted FEs and to limit the size of the ensemble, experiences with fewer than five classes are not considered. Second, the addition of FEs to the ensemble is stopped after 85\% of the classes have been seen, because once robust features for most classes have been learned, good performance on the remaining ones is expected. Additionally, an FE is always trained on the first experience. With these constraints, the proposed approach learns feature extractors based on the highlighted experiences shown in Figure~\ref{fig:horde-heuristic}.

\textbf{Feature extractor training:} When an FE is trained on the current experience, learning occurs with the usual cross-entropy loss on a fully connected head with as many outputs as classes. In order to promote the learning of features in a more similarly distributed space, a contrastive loss with emphasis on the hard negative pairs is also included on a separate head, as shown in Figure~\ref{fig:horde}a. Both losses are balanced with an adaptive alpha, which is automatically computed based on the energy of each loss. After learning that experience, the heads are removed and the backbone is frozen and added to the rest of the ensemble.

\textbf{Pseudo-feature projection:} Regardless of an FE being trained and added to the ensemble for the current experience, the unified head that uses all the representations is always trained from all FEs and learns to discriminate between all seen classes (Figure~\ref{fig:horde}b). The proposed pseudo-feature projection extends feature translation~\cite{petit2023fetril} with the corresponding class standard deviation to allow a better sampling of the dimensions. It translates a representation from a class into a projected representation of a different class. Let $a_i$ be the concatenated representation of all FEs outputs for a sample from the current experience belonging to class~$i$. The proposed projection is then defined as
\[
\hat{a}_{j, i} = \mu_j + \frac{a_{i} - \mu_i}{\sigma_i} \cdot \sigma_j \text{,}
\]
where $\hat{a}_{j, i}$ is the estimated projection from class $i$ to class $j$. This projection is applied to each sample in the training batch while learning the unified head, and not used during evaluation. The target class~$j$ is chosen at random from previously learned classes, and both the original representation and the projected one are added onto the loss.
The class prototypes (i.e.,~the mean~$\mu_{i}$ and standard deviation~$\sigma_{i}$) are always updated before the unified head is trained by calculating the statistics over the available class data. However, since multiple FEs exist, access might not be available to the mean and standard deviation for each class, depending on when they were learned, or if those classes have appeared since that time. Therefore, if the class statistics are unknown for a feature extractor $e$, estimates are needed for $\hat{\mu}_{c, e}$ and $\hat{\sigma}_{c, e}$. We fix the estimation of the standard deviation to $\hat{\sigma} = 1$, as we do not have any information about the class shape for this feature extractor. For the estimation of the unknown mean statistic of a specific FE output, we resort to the simple heuristic of using the original representation of the current sample without modification: $\hat{\mu}_{c, e} = a_{i, e}$.

Pseudocode for Horde as well as further experimental analysis is presented in Appendix \ref{sec:appendix_mmasana}. The implementation is publicly available on GitHub: \href{https://github.com/mmasana/clvision-chlg-2023/}{https://github.com/mmasana/clvision-chlg-2023/}.

\section{Strategy 3: DWGRNet}
\label{sec:pddbend}

The strategy proposed by team \emph{pddbend} is called  \textbf{D}ynamic \textbf{W}eighted \textbf{G}ated \textbf{R}epresentation \textbf{Net}work (\textbf{DWGRNet}). This strategy creates independent branches for each experience, and uses gating units to control which branches are active. During training, the branch corresponding to the current experience is activated by its gating unit, facilitating learning, while branches from previous experiences remain inactive. To improve the model's generalization and robustness, data augmentation techniques, such as AugMix \cite{hendrycks2019augmix}, are used to enhance the data samples.
In the testing phase, the gating units control the sequence of predictions. Importantly, using predictions from all branches, as done in~\cite{yan2021dynamically}, may not always yield optimal results. This is because many branches may not have been exposed to a particular class, leading to overconfident and potentially inaccurate predictions. This issue resembles an open-set recognition problem.
To mitigate this, DWGRNet assigns weights based on entropy, feature norm, and the number of classes experienced by each branch. Specifically, the entropy of each branch's predicted probability distribution is assessed. High entropy indicates the possibility of the sample being an open-set item for that branch. Similarly, the feature norm is computed. A higher feature norm suggests that the sample is likely to be an open-set sample. Finally, it is posited that experiences with a larger number of classes will render the model's prediction more reliable. As a result, weighting can also adjusted based on the number of classes in each experience.

\subsection{Motivation and Related Work}

To achieve a good balance between stability and plasticity in class-incremental learning, reference~\cite{yan2021dynamically} introduced the method Dynamically Expandable Representation (DER). DER decouples the adaptivity of feature representation from the classification procedure. When faced with a new task that contains novel categories, the DER method first freezes the previously learned feature extractor. Then, it incorporates a new feature extractor to expand the main feature extractor network. Ultimately, the features extracted by all the extractors are combined and forwarded to the classifier module for predicting the category. This strategy exhibits notable efficiency in preserving existing knowledge while providing sufficient flexibility to capture new information.

However, the DER architecture cannot be directly employed in the context of this competition. This is because DER constructs a model ensemble that exploits replay buffers for raw sample storage. Another drawback of DER is the simultaneous execution of predictions across all branches, which has high computational complexity. In practice, DER applies predictions from each branch without distinction. A problem with this is that certain branches produce overconfident predictions for unseen classes, leading to sub-optimal performance.

To fix the limitations of DER, DWGRNet implements an open-set approach \cite{geng2020recent,Sun_Guo_Li_2021,vaze2022openset}. Within this framework, each model is restricted to be exposed to a limited number of classes. This approach is inspired by recent advancements in open-set recognition literature that have elucidated the relationships between logit entropy, feature norm, and out-of-distribution (OOD) samples. Capitalizing on these findings, a scoring system is designed for OOD samples.
This scoring system requires the evaluation of each model to determine whether a test sample falls within its OOD category, prior to the ensembling process. This evaluation allows two possible options: (1) to either suppress the OOD outputs or (2) to harness the OOD scores to assign weight to each sample. Therefore, it serves to alleviate the constraints of the DER method, with the overarching aim of augmenting both its performance and applicability in the realm of CIR.

\subsection{Method Description}

\begin{figure}[t]
\centering
\includegraphics[width=0.99\textwidth]{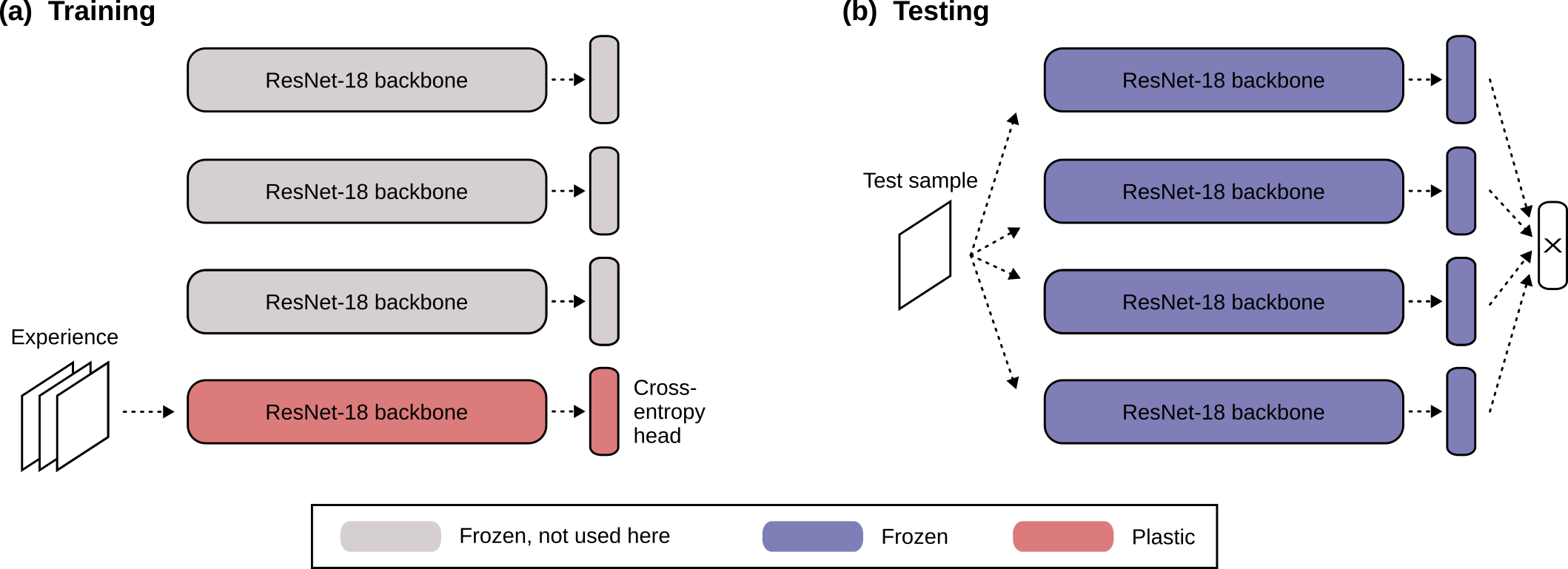}
\vspace{0.2cm}
\caption{\revision{Schematic of DWGRNet during training and test time. \textbf{(a)} When training on a new experience, a new model branch is trained on that experience, after which that branch is frozen and added to the model ensemble. \textbf{(b)} At test time, a test sample is forwarded through all model branches. The outputs of each branch are collected and combined according to equation~(\ref{eq:dwgrnet}) for the final prediction.}}
\vspace{-0.2cm}
\label{fig:dwgrnet}
\end{figure}

DWGRNet uses gating units to control the activation of each branch. A new branch is added with each new experience and then activated, while the model parameters in the old branches are kept frozen. As illustrated in Figure~\ref{fig:dwgrnet}a, during the training phase, no special loss functions or replay buffers are used. Instead, the standard cross-entropy loss is used to train the model, while using AugMix to enhance the model's generalization and robustness. AugMix combines different data augmentation techniques. During the testing phase, the gating units could activate each branch one by one to avoid the need for a large GPU memory. Their outputs are collected and then used to make the final prediction. This process is illustrated in Figure~\ref{fig:dwgrnet}b.

This method is similar to a model ensemble. It is designed in a way that both complies with the competition rules and also speeds up training time and reduces memory space during training and inference. On the other hand, the ``naive'' model ensembling approach may not work properly as each model only learns limited knowledge. Therefore, the open-set idea is employed. If the problem is treated as an open-set problem, and asks each model whether a test sample is an OOD sample before ensembling, it is possible to either mask the OOD outputs or use the OOD scores to weigh each sample.

The method relies on the concept of entropy to make the strategy work. The assumption is that the entropy of a branch's outputs can indicate their confidence. For an in-distribution sample, the branch's outputs should have low entropy since the branch has seen the class and is confident in its prediction. For an OOD sample, the branch's outputs should have high entropy since the branch has not seen this class before, and lacks confidence in its output. While this phenomenon is not always accurate, it holds true in many circumstances.

Additionally, the aim is to further enhance the performance of the model ensemble. First, the reliability of each model branch is to be evaluated. The hypothesis is that the more classes a branch encounters, the more reliable it becomes. Therefore, the number of classes ($N_C$) is used to weight each branch's output. Finally, inspired by the idea of ``a good closed-set classifier is all you need'' in \cite{vaze2022openset},  feature norms are used to verify whether the branch perceives the sample as an in-distribution or OOD sample. The OOD samples tend to have a lower feature norm.

To calculate the final logits, the $k$-th output logit of the model ensemble is calculated as:
\begin{equation}
    \text{ensemble\_logit}_k = \max_{i \in M}{(\frac{\text{logit}_{i,k}}{\text{entropy}_{i,k}} * {N_C}^{(i)} * \text{feature\_norm}_{i,k})},
    \label{eq:dwgrnet}
\end{equation}
where $M$ is the set of saved model branches. An ablation study to investigate the role of each module can be found in Appendix \ref{sec:appendix_pddbend}.

\section{Results}
\label{sec:results}

\subsection{Pre-selection Phase}

In the pre-selection phase of the challenge, the teams were ranked based on their average accuracy. The average accuracy for each team was computed as the mean accuracy on the test set after training on each of the three streams released for the first phase, namely $S_1$, $S_2$, and $S_3$. The top ten teams in the pre-selection phase are shown in Table \ref{tab:preselection_results}. The top five teams proceeded to the final phase.

\begin{table}[H]
    \centering
    \begin{tabular}{l c c c c} \toprule
         \multicolumn{1}{c}{Team} & Average Accuracy  $\uparrow$  & Accuracy $S_1$ & Accuracy $S_2$ & Accuracy $S_3$  \\
         \cmidrule(lr){1-1} \cmidrule(lr){2-5} 
         1. pddbend & $44.77 $ &  $52.32 $ & $40.31 $ & $41.67 $ \\
         2. linzz	& $44.08 $ & 	$50.88$ &	$41.83$ & $39.52$ \\
         3. shelley	& $42.53 $  &  $45.21 $ &	$41.80 $ & $40.58 $ \\
         4.	xduan7	& $41.37 $ &   $44.99 $ &	$40.80 $ &	$38.31 $ \\
         5. mmasana	& $40.52 $ &	$42.47 $ & $34.29 $ & $44.81 $ \\
         \cmidrule(lr){2-5}
         6. zys\_tjut	& $34.35 $ &   $45.34 $ &	$30.43 $ &	$27.28 $  \\
         7. Vanixxz	& $31.72 $ &	$33.33 $ &	$30.01 $ &	$31.83 $ \\
         8. M1andy	& $31.01 $ & 	$33.32 $ &	$28.97 $ &	$30.74 $ \\
         9. rainstar	& $30.88 $ &	$36.01 $ &	$27.77 $ & 	$28.87 $ \\
         10. flyfishzzz	& $29.97 $ & 	$31.31 $ & 	$31.31 $ &	$27.28 $ \\
         
         
    \bottomrule
    \end{tabular}
    \vspace{0.2 cm}
    \caption{Results from the pre-selection phase for the top ten teams. Shown is the accuracy~(as~\%) on the test set of CIFAR-100 after training on each of the three streams from the pre-selection phase.}
    \label{tab:preselection_results}
\end{table}

\subsection{Final Phase}

In the final phase of the challenge, three streams, namely $S_4$, $S_5$, and $S_6$, with configurations different from those in the pre-selection phase streams, were used to evaluate the solutions submitted by the finalist teams. The details of the challenge stream configurations are given in Table \ref{tab:stream_configurations} in the Appendix \ref{sec:stream_configurations}. The final phase demonstrated substantial variations in performance among the solutions of the finalist teams, as presented in Table~\ref{tab:final_results}.  Team \textit{xduan7} (Section~\ref{sec:xduan}) was selected as the challenge winner with an average accuracy of \(62.75\%\), thereby substantially outperforming the other finalists. 
Team \textit{linzz} earned the second spot with an average accuracy of \(45.02\%\). However, they decided not to contribute their solution to this report. The other two teams, \textit{mmasana} (Section~\ref{sec:mmasana}) and \textit{pddbend} (Section~\ref{sec:pddbend}), achieved average accuracies of \(41.11\%\) and \(40.91\%\), respectively. Recorded video presentation of the submitted solutions by the finalist teams can be accessed through \href{https://sites.google.com/view/clvision2023/challenge/challenge-results}{https://sites.google.com/view/clvision2023/challenge/challenge-results}. 

\revision{\paragraph{Baselines}
To put the performance of the solutions of the finalist teams in perspective, we also report in Table~\ref{tab:final_results} the performance of several popular CL baselines strategies after training on each of the three data streams from the final phase.
The approach `Naive' refers to sequential fine-tuning the Slim-ResNet-18 backbone without any continual learning mechanism. For Elastic Weight Consolidation~(EWC)~\cite{kirkpatrick2017overcoming}, a popular parameter regularization method, the lambda hyperparameter is set to $1$; using higher or lower lambda values results in slightly lower average accuracy. For Learning without Forgetting~(LwF)~\cite{li2017learning}, a popular functional regularization method, the alpha and temperature hyperparameters are set to their default values of $1$ and $2$, respectively. For Experience Replay~(ER), two versions are run: one with a memory buffer with total capacity of $200$~samples and another with total capacity of $2000$~samples. For all baselines the implementation of the Avalanche library (version 0.3.1) is used. For `Joint', the backbone model is trained in an \textit{iid} fashion with access to samples from all classes at the same time.

The comparison with the CL baseline strategies shows that there is a significant gap between the performance of the finalist solutions and these baselines. In particular, even ER with a moderately-sized buffer (i.e.,~2000~samples) performs substantially worse than all of the finalist solutions. Notable is that the solution of team~\textit{xduan7} approaches the performance of jointly training the backbone model on all training data at the same time.}

\begin{table}[H]
    \centering
    \begin{tabular}{l c c c c l} 
    \toprule
         \multicolumn{1}{c}{Team} & Average Accuracy $\uparrow$ & Accuracy $S_4$ & Accuracy $S_5$ & Accuracy $S_6$  \\
         \cmidrule(lr){1-1} \cmidrule{2-5}
         
         1. xduan7 & $62.75$  & $63.64$	& $68.04$ &	$56.57$	\\
         2. linzz	& $45.02$ & $46.21$ &	$47.27$ & $41.58$	 \\
         3.	mmasana	& $41.11$ & $40.82$	& $45.79$ &	$36.72$	\\
         4. pddbend	& $40.91$ & $42.07$	& $44.44$ &	$36.23$	 \\
    \midrule
    \revision{Naive} & \revision{$~7.83$} & \revision{$~9.11$}	& \revision{$11.48$} &	\revision{$~4.83$}	 \\
    \revision{EWC}	& \revision{$~7.14$} & \revision{$~8.28$}	& \revision{$~8.67$} &	\revision{$~4.49$}	 \\
    \revision{LwF}	& \revision{$~9.83$} & \revision{$10.17$}	& \revision{$13.46$} &	\revision{$~5.85$}	 \\ 
    \revision{ER (buffer: 200)}	& \revision{$~9.87$} & \revision{$~8.98$}	& \revision{$13.48$} &	\revision{$~7.17$}	 \\ 
    \revision{ER (buffer: 2000)}	& \revision{$21.91$} & \revision{$18.56$}	& \revision{$27.58$} &	\revision{$19.63$}	 \\ 
    \midrule
    \revision{Joint}	& \revision{$65.12$} & \revision{$-$}	& \revision{$-$} &	\revision{$-$}	 \\ 
    \bottomrule     
    \end{tabular}
    \vspace{0.2 cm}
    \caption{Results from the final phase for the finalist teams,  \revision{along with the performance of popular CL baselines}. Team \textit{shelley} is not included since they decided not to participate in the final phase after realizing their solution violated the challenge rules.}
    \label{tab:final_results}
\end{table}

\revision{\subsection{Additional Experiments}

\subsubsection{No Repetition}
To test whether the repetition in the data stream plays an important role in the effectiveness of the finalist solutions, they are also evaluated after training on a data stream without repetition. For this, the ``standard'' CIFAR-100 class-incremental learning benchmark is used, with 20 experiences (or tasks) that are encountered one after the other, whereby each experience contains five distinct classes. The results in Table~\ref{tab:results_no_repetition} indicate that the performance of the finalist solutions is significantly reduced when there is no repetition in the data stream.
Importantly, while ER with buffer size of 2000 was clearly outperformed by each of the finalist solutions on the data streams with repetition, when there is no repetition, this version of ER performs better than all the finalist solutions. This shows that repetition in the data stream can change the relative effectiveness of different CL strategies.}
\begin{table}[H]
    \centering
    \begin{tabular}{l c } 
    \toprule
         \multicolumn{1}{c}{\revision{Team}} & \revision{Accuracy $\uparrow$}  \\
         \cmidrule(lr){1-1} \cmidrule{2-2}
         
         \revision{xduan7} & \revision{$24.30$} \\
         \revision{mmasana}	& \revision{$~3.39$} 	\\
         \revision{pddbend}	& \revision{$~7.59$} \\
    \midrule
    \revision{Naive} & \revision{$~1.67$} \\
    \revision{EWC} & \revision{$~1.71$} \\
    \revision{LwF} & \revision{$~1.82$} \\
    \revision{ER (buffer: 200)} & \revision{$~3.74$} \\
    \revision{ER (buffer: 2000)} & \revision{$25.35$} \\
    \midrule
    \revision{Joint} & \revision{$65.12$} \\
    
    \bottomrule     
    \end{tabular}
    \vspace{0.2 cm}
    \caption{\revision{Results for the class-incremental learning experiments on CIFAR-100 without repetition. Shown is the accuracy~(as~\%) on the test set at the end of training. Team \textit{linzz} is not included in these additional experiments as they decided not to participate in the writing of this report.}}
    \label{tab:results_no_repetition}
\end{table}

\revision{\subsubsection{Tiny ImageNet}
To probe the generalizability of our results, in this section the finalist solutions are evaluated on three CIR data streams generated using the Tiny ImageNet dataset~\cite{le2015tiny}. This dataset contains natural images of $64\times64$~pixels, divided over $200$~classes, of which we only use a random subset of $100$~classes. The configurations to generate the data streams are the same as those used for the data streams $S_4$, $S_5$ and $S_6$ in the final phase of the challenge. The evaluation results are shown in Table~\ref{tab:results_tin}. These results indicate consistency in the performance of the three finalist solutions relative to the dataset used to generate the data streams.}

\begin{table}[H]
    \centering
    \begin{tabular}{l c c c c l} 
    \toprule
         \multicolumn{1}{c}{\revision{Team}} & \revision{Average Accuracy $\uparrow$} & \revision{Accuracy $S_4^{\text{TIN}}$} & \revision{Accuracy $S_5^{\text{TIN}}$} & \revision{Accuracy $S_6^{\text{TIN}}$}  \\
         \cmidrule(lr){1-1} \cmidrule{2-5}
         
         \revision{xduan7} & \revision{$55.62$}  & \revision{$55.52$}	& \revision{$56.48$} &	\revision{$54.88$}	\\
         \revision{mmasana}	& \revision{$34.01$} & \revision{$34.94$}	& \revision{$38.14$} &	\revision{$28.94$}	\\
         \revision{pddbend}	& \revision{$32.68$} & \revision{$36.66$}	& \revision{$39.12$} &	\revision{$22.28$}	 \\
    \bottomrule     
    \end{tabular}
    \vspace{0.2 cm}
    \caption{\revision{Results for the finalist solutions after training them on three CIR data streams constructed from the Tiny ImageNet dataset. Team \textit{linzz} is not included in these additional experiments since they decided not to participate in the writing of this report.}}
    \label{tab:results_tin}
\end{table}

\section{Discussion}
\label{sec:discussion}

The challenge of class-incremental learning with repetition asks for a careful balance between learning new information and not forgetting previous knowledge. This report has described three solutions proposed by the finalists of the CLVision challenge at CVPR~2023. These solutions reflect diverse methodologies to address the challenge of \textit{learning continually in the presence of repetition}. Interestingly, an approach that is shared by all finalist solutions is an ensemble-based strategy. This commonality suggests that the relatively simple approach of having separate (sub-)networks per experience might, from a practical perspective, be a very useful strategy for continual learning\revision{, especially when there is repetition in the data stream}.

The suitability of the ensemble-based strategy for this challenge is illustrated well by the winning solution~\mbox{\emph{HAT-CIR}}. During the pre-selection phase, the solution submitted by this team used a single network with a HAT-based partitioning approach, which can be interpreted as a relatively soft version of an ensemble-based strategy. In the final phase of the challenge, this team switched to using network replicas, a more explicit version of an ensemble-based strategy, which substantially improved the performance.
The second solution discussed in this report, \textit{Horde}, trained an ensemble of feature extractors, where a new feature extractor was added only when needed based on a heuristic. It also used a unified head to discriminate between classes that are observed in the current step of training. The third solution, \textit{DWGRNet}, used independent branches for each experience, activated by gating units during training and incorporating data augmentation techniques for robustness. This solution further approached the problem from an open-set problem perspective, and to address open-set recognition issues, it employed a weighting strategy based on factors such as entropy, feature norm, and the number of classes in each experience.

\revision{In recent years, several variants of ensemble-based strategies have been explored for continual learning. Two broad types of ensemble-based strategies can be distinguished. On the one hand, a number of studies have explored using an ensemble of multiple models that are all continually trained on the full data stream~\cite{wang2022coscl,wang2023incorporating,doan2023continual,rypesc2024divide}. On the other hand, it is also possible to have an ensemble of multiple models whereby each model is trained on a different part of the data stream (e.g.,~to have a separate model for each task or experience)~\cite{vogelstein2020general,yan2021dynamically,hess2023knowledge}. 
In this challenge, the finalist teams predominantly used the second type of ensemble-based strategy, although the \emph{ensembles} used by HAT-CIR are an example of the first type (while the \emph{fragments} of HAT-CIR belong to the second type).}

\revision{The results reported in this paper suggest that when there is repetition in the data stream, the ensemble-based strategy might be particularly effective for continual learning. Indeed, we found that while the ensemble-based solutions of the finalists teams clearly outperformed all baseline methods on the data streams with repetition, when the repetition was removed, the finalist solutions became markedly less effective, and they were outperformed by experience replay with a moderately-sized memory buffer. This thus establishes that repetition in the data stream affects the behaviour of different CL strategies in different ways. An exciting question for future work is to map out exactly how the amount and the type of repetition affect the effectiveness of each approach.}

\revision{Another intriguing question is why the ensemble-based approach is so effective for continual learning in the presence of repetition. We speculate that one reason is that} the repetition allows each class to be observed in multiple experiences and in different pairings with other classes. This diversity of observed combinations of classes might allow for the development of a ``rich'' set of feature extractors, leading to more robust representations. \revision{Another perspective is that the repetition in the data stream can be interpreted as providing a natural form of bagging, on which an ensemble method can be effectively trained.}

Overall, the solutions proposed by the winning teams of the CLVision challenge at CVPR~2023 offer diverse yet converging approaches for the problem of class-incremental learning with repetition. While the proposed solutions are primarily based on the idea of \textit{ensemble learning}, they also leverage specialized discrimination techniques, data augmentation and weighting strategies. This report provides multiple avenues for future research to further refine and combine these approaches for more effective and robust class-incremental learning strategies, especially when repetition is present in the data stream.

\printcredits

\section*{Declaration of Generative AI and AI-assisted technologies in the writing process}
During the preparation of this work, some of the authors used ChatGPT (GPT-4) in order to improve the readability and language of certain sections. After using this tool, the authors reviewed and edited the content as needed and take full responsibility for the content of the publication.

\section*{Acknowledgements}
This work was supported by the Exascale Computing Project, a collaborative effort of the U.S. Department of Energy Office of Science and the National Nuclear Security Administration [grant number 17-SC-20-SC]; by TU-Graz SAL DES Lab, part of the ``University SAL Labs'' initiative of Silicon Austria Labs (SAL) and its Austrian partner universities for applied fundamental research for electronic based systems; by the National Science and Technology Major Project [grant number 2022ZD0114801]; by a senior postdoctoral fellowship from the Research Foundation -- Flanders (FWO) [grant number 1266823N]; and by a Marie Skłodowska-Curie fellowship under the European Union's Horizon Europe programme [grant number 101067759].

\bibliographystyle{cas-model2-names}

\bibliography{main}

\bio{}
\endbio

\newpage
\appendix

\renewcommand\thefigure{\thesection.\arabic{figure}}  
\renewcommand\thetable{\thesection.\arabic{table}}

\section{Author Contributions}\label{sec:author_contributions}

In addition to the CRediT authorship contribution statement, here we provide additional information regarding to which authors have been responsible for the development and description of the different strategies reported on in this paper:
\begin{itemize}
    \item \textbf{Strategy 1: HAT-CIR} developed by Xiaotian Duan, Zixuan Zhao, and Fangfang Xia. 
    \item \textbf{Strategy 2: Horde} developed by Marc Masana, Benedikt Tscheschner, and Eduardo Veas.
    \item \textbf{Strategy 3: DWGRNet} developed by Yuxiang Zheng, Shiji Zhao, Shao-Yuan Li, and Sheng-Jun Huang.
\end{itemize}

\section{Participation Over Time}\label{sec:participation_over_time}
\setcounter{figure}{0}
\setcounter{table}{0}

\begin{figure}[!htb]
    \centering
    \subfloat[Highest average accuracy per day.]{
        \includegraphics[width=0.42\textwidth]{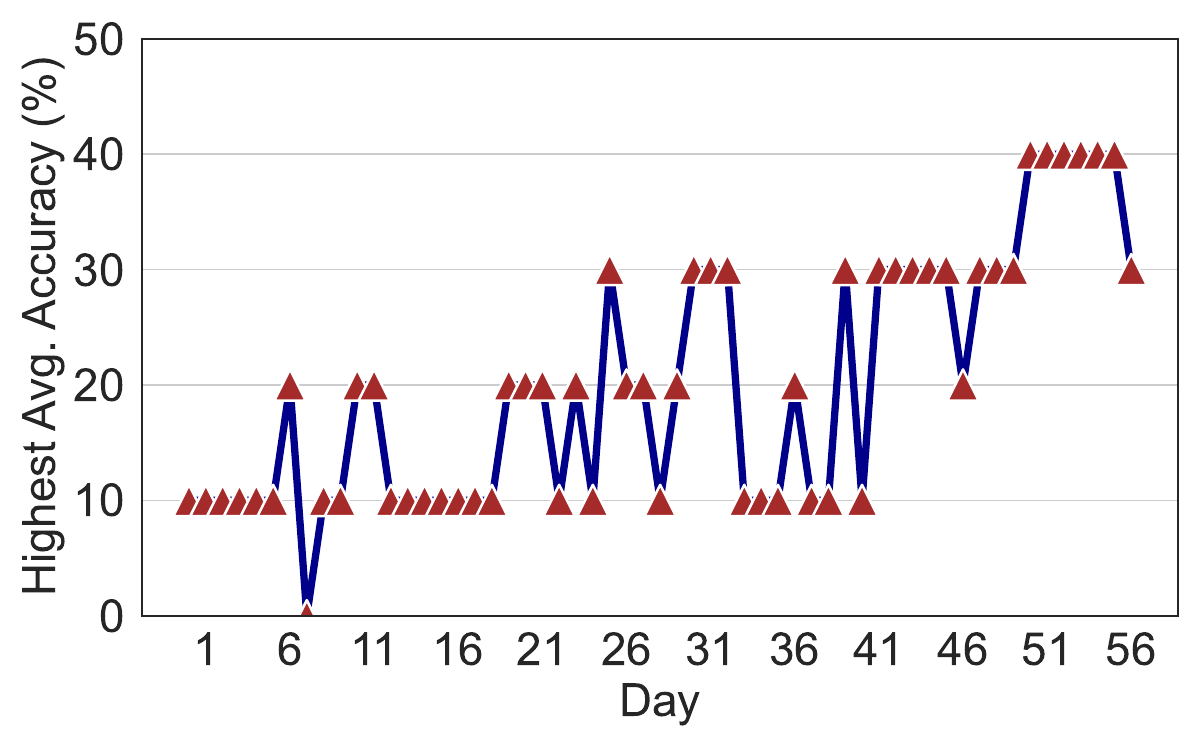}
        \label{fig:subfigA1}
    }
    \subfloat[Number of submissions per day.]{
        \includegraphics[width=0.42\textwidth]{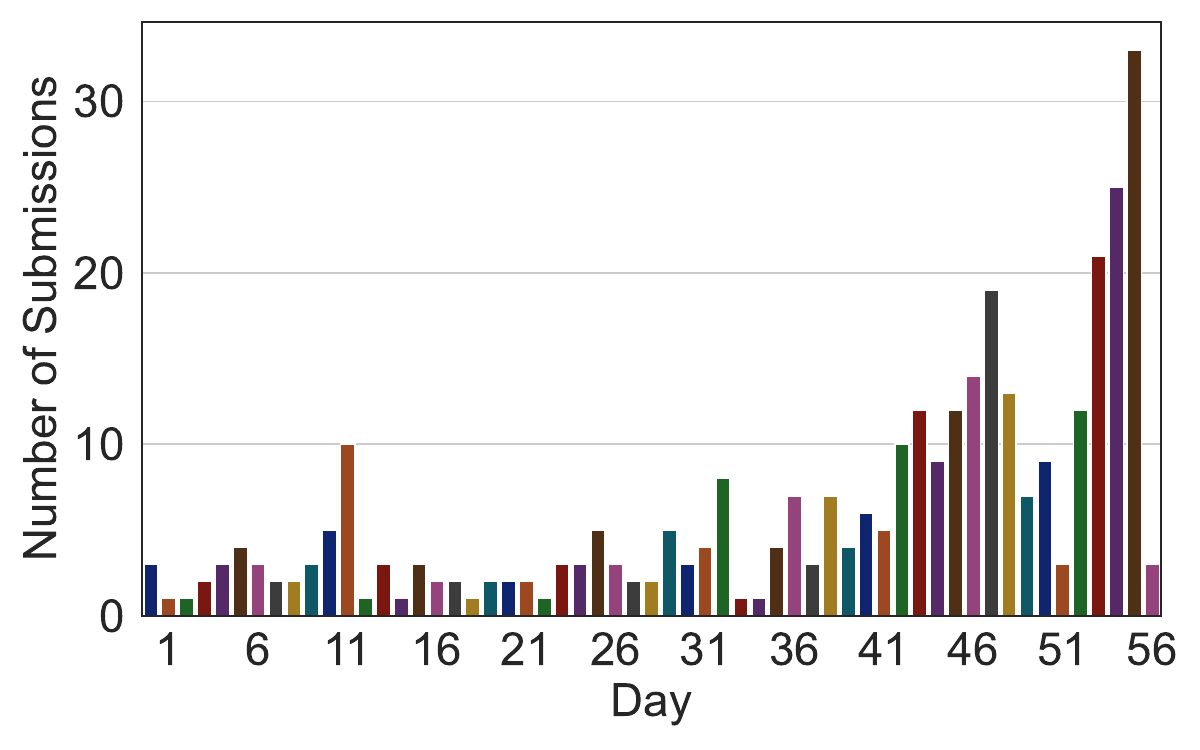}
        \label{fig:subfigA2}
    }

    \caption{Participation over time in the pre-selection phase. Accuracy values are rounded by the competition platform.}
    \label{fig:participation_over_time}
\end{figure}

The graphs in Figure \ref{fig:participation_over_time} show the number of submissions per day by all teams, and the maximum average accuracy achieved on each day. In summary, participants began with a low average accuracy of $\sim10\%$. Over time, the highest average accuracy showed gradual improvement, peaking and stabilizing at $\sim40\%$ towards the end. The challenge saw a steady increase in engagement, with daily submissions starting from as low as one to three in the beginning, and reaching above $30$ towards the end.

\section{Stream Configurations}\label{sec:stream_configurations}

The configurations of the streams used for both pre-selection and final phases are given in Table \ref{tab:stream_configurations}.

\begin{table}[h]
\centering
\begin{tabular}{c c c}
\toprule
\text{Stream} & First Occurrence $P_f$ & \text{Repetition Dist. $P_r$} \\
\cmidrule(lr){1-1} \cmidrule(lr){2-3} 
$S1$ & Geometric: $p=0.5$ & Zipfian: $e=0.7$  \\ 
$S2$ & Geometric: $p=0.001$ & Zipfian: $e=0.8$  \\ 
$S3$ & Geometric: $p=0.2$ & Fixed at $0.04$  \\ 
\midrule
$S4$ & Geometric: $p=0.6$ & Zipfian: $e=0.8$  \\ 
$S5$ & Geometric: $p=0.001$ & Zipfian: $e=0.6$  \\ 
$S6$ & Geometric: $p=0.1$ & Fixed at $0.05$  \\ 
\bottomrule
\end{tabular}
\vspace{0.2cm}
\caption{Configurations of the challenge streams.}

\label{tab:stream_configurations}
\end{table}

\section{Appendix -- HAT-CIR} \label{sec:appendix_xudan}
\setcounter{figure}{0}
\setcounter{table}{0}

\subsection{Ablation Study for Single-model Settings}

The ablation study, presented in Table~\ref{tab:hat-cir-ablation}, indicates that various components of the proposed method were important for achieving robust performance in the pre-selection phase, when network replicas were not yet used. Notably, the modified HAT-CL technique significantly outperforms the traditional HAT method in the context of CIR, thus confirming its efficacy. The supervised contrastive learning phase also plays a crucial role in learning better feature representations, as its absence led to a marked decline in performance.

\begin{table}[h]
\centering
\begin{tabular}{c c c c c}
\toprule
\text{Method} & \text{Accuracy $S_4$} & \text{Accuracy $S_5$} & \text{Accuracy $S_6$} & \text{Avg. Decrease from Baseline} \\
\cmidrule(lr){1-1} \cmidrule(lr){2-5} 
Baseline & 39.08\% & 42.89\% & 34.59\% & - \\  
With Original HAT & 23.89\% & 26.05\% & 19.47\% & 15.71\% \\  
Without SupCon & 35.39\% & 28.93\% & 22.16\% & 10.02\% \\  
Without Momentum \& TTA & 30.67\% & 35.16\% & 26.61\% & ~8.04\% \\  
\bottomrule
\end{tabular}
\vspace{0.2cm}
\caption{Ablation study results for HAT-CIR when using a single model without using network replicas. Shown is the test accuracy~(as~\%) after training on each data stream from the final phase of the challenge. The ``Avg.\ Decrease from Baseline'' column indicates how much performance is lost relative to the baseline (i.e., the version of HAT-CIR that was used in the pre-selection phase of the challenge). The term ``Original HAT'' refers to the original hard attention method, ``SupCon'' stands for supervised contrastive learning, and ``Momentum \& TTA'' are abbreviations for momentum-based test-time decision-making and test-time augmentation, respectively.}
\label{tab:hat-cir-ablation}
\end{table}

\subsection{Influence of Number of Network Replicas}

The results presented in Table~\ref{tab:hat-cir-replicas} demonstrate that both increases in ensemble replicas and increases in fragment replicas enhance performance. Yet, by adding more replicas, the added benefit starts to diminish. This suggests there is a sweet spot in balancing the number of fragments and ensembles for optimal performance without excessive training and/or test time. Importantly, adding ensemble replicas comes with the trade-off of both longer training and test time, while adding fragment replicas only comes with the trade-off of longer test time. Because the rules of the challenge predominantly put restrictions on computational resources during training, the final version of the model that was selected predominantly used fragment replicas.

\begin{table}[h]
\centering
\begin{tabular}{c c c  c c c}
\toprule
\text{\# of Replicas} & \text{\# of Fragments} & \text{\# of Ensembles}  & \text{Accuracy $S_4$} & \text{Accuracy $S_5$} & \text{Accuracy $S_6$} \\
\cmidrule(lr){1-1} \cmidrule(lr){2-3} \cmidrule{4-6}
1 & 1 & 1 & 39.08\% & 42.89\% & 34.59\% \\
2 & 2 & 1 & 40.20\% & 43.03\% & 35.90\% \\
2 & 1 & 2 & 43.90\% & 46.17\% & 39.11\% \\
10 & 5 & 2 & 45.96\% & 49.42\% & 42.18\% \\
10 & 2 & 5 & 48.06\% & 48.50\% & 44.69\% \\
50 & 10 & 5 & 51.14\% & 53.22\% & 46.37\% \\
50 & 5 & 10 & 50.70\% & 52.58\% & 47.92\% \\
100 & 10 & 10 & 52.74\% & 54.13\% & 46.87\% \\
\cmidrule(lr){1-1} \cmidrule(lr){2-3} \cmidrule{4-6}
$\mathbf{100}$ & $\mathbf{50}$ & $\mathbf{2}$ & $\mathbf{63.64}\%$ & $\mathbf{68.04}\%$ & $\mathbf{56.57}\%$ \\
\bottomrule
\end{tabular}
\vspace{0.2cm}
\caption{Effect of the number and types of network replicas on the performance of HAT-CIR on the data streams from the final phase of the challenge. The table shows the test accuracy~(as~\%) achieved for each data stream with different numbers of fragments and ensembles.}
\label{tab:hat-cir-replicas}
\end{table}

\section{Appendix -- Horde}\label{sec:appendix_mmasana}
\setcounter{figure}{0}
\setcounter{table}{0}

\subsection{Algorithm}

Inspired by zero-forgetting methods, Horde promotes stability on each task-specific Feature Extractor (FE) while leveraging class repetition for balancing plasticity by learning an aligned unified representation. Each extractor provides a discriminative representation for the seen classes of the task where it is learned. Then, pseudo-feature projection is used to achieve alignment on the unified head. A description of the pseudo-code of Horde is provided in Algorithm~\ref{alg:horde-unified}. Additionally, the feature extractor training steps are shown in Algorithm~\ref{alg:horde-fe-training}.

\begin{minipage}{0.48\textwidth}
\begin{algorithm}[H]
    \centering
    \caption{Horde Algorithm}
    \label{alg:horde-unified}
    \begin{algorithmic}[1] 
        \For{experience}
            \If{Training Condition} 
                \State Train FE and add to ensemble \Comment{See Alg. \ref{alg:horde-fe-training}}
            \EndIf
            \State Calculate $\mathbf{\mu}_c$ and $\mathbf{\sigma}_c$ for all classes $c$
            \For{training epoch} \Comment{Only Unified head}
                \For{Batch}
                    \State Generate $\hat{a}_{j, i}$ from $a_i$ and old classes
                    \State Calculate $\mathcal{L}_{CE}$ with both $a_{i}$ and $\hat{a}_{j, i}$
                    \State Backprop $\mathcal{L}_{CE}$
                \EndFor
            \EndFor
        \EndFor
    \end{algorithmic}
\end{algorithm}
\end{minipage}
\hfill
\begin{minipage}{0.48\textwidth}
\begin{algorithm}[H]
    \centering
    \caption{FE Training}
    \label{alg:horde-fe-training}
    \begin{algorithmic}[1] 
        \State Initialize CE and ML Head
        \State Initialize new or transfer learning model
        \For{training epoch}
            \For{Batch}
                \State Calculate $\mathcal{L}_{CE}$
                \State Calculate $\mathcal{L}_{ML}$
                \State Backprop $(1 - \alpha) \cdot \mathcal{L}_{CE} + \alpha \cdot \mathcal{L}_{ML}$
            \EndFor
            \State Update $\alpha$ from avg. $\mathcal{L}_{CE}$ and $\mathcal{L}_{ML}$
        \EndFor
        \State Remove CE and ML head
        \State Freeze FE
    \end{algorithmic}
\end{algorithm}
\vspace{0.01cm}
\end{minipage}

\subsection{Additional Experimental Analysis}
\begin{figure}[h]
    \centering
    \includegraphics[trim= 10 8 20 8, clip, width=0.32\textwidth]{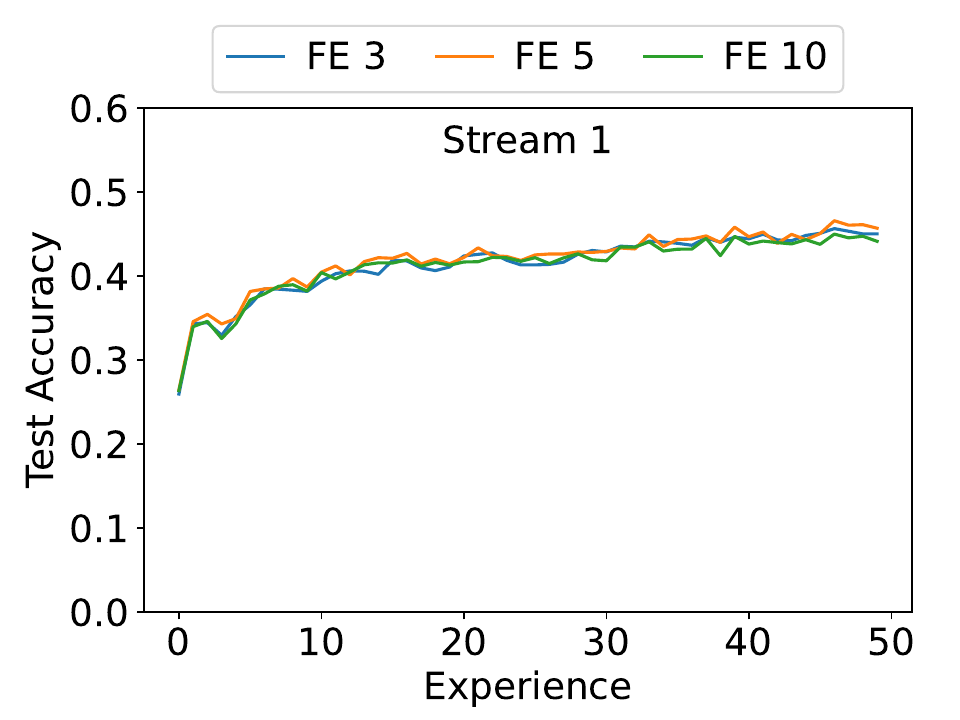}
    \includegraphics[trim= 10 8 20 8, clip, width=0.32\textwidth]{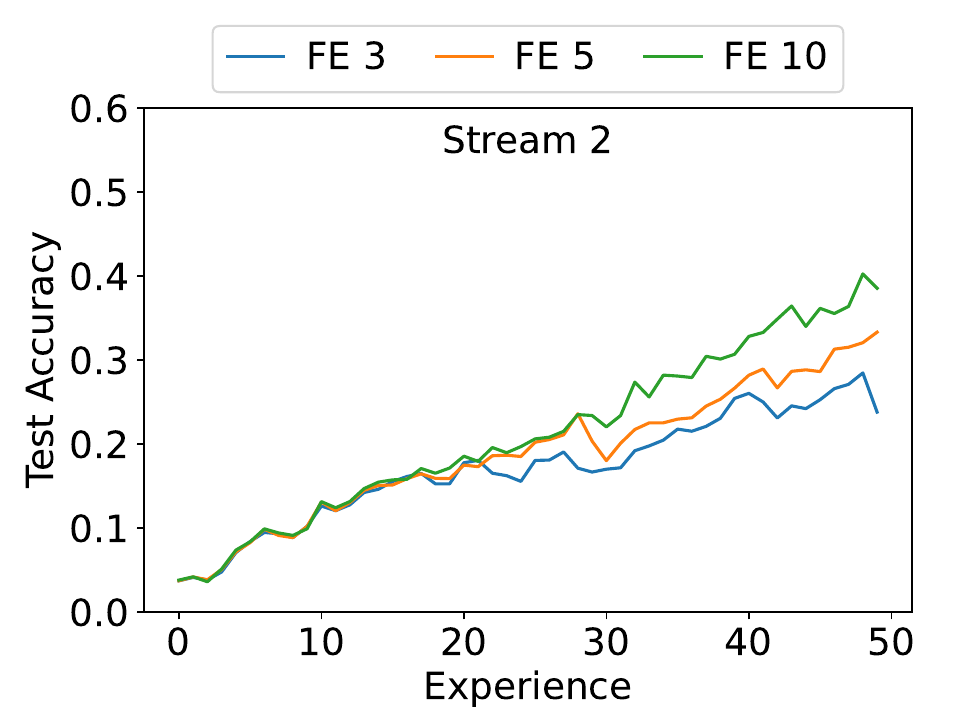}
    \includegraphics[trim= 10 8 20 8, clip, width=0.32\textwidth]{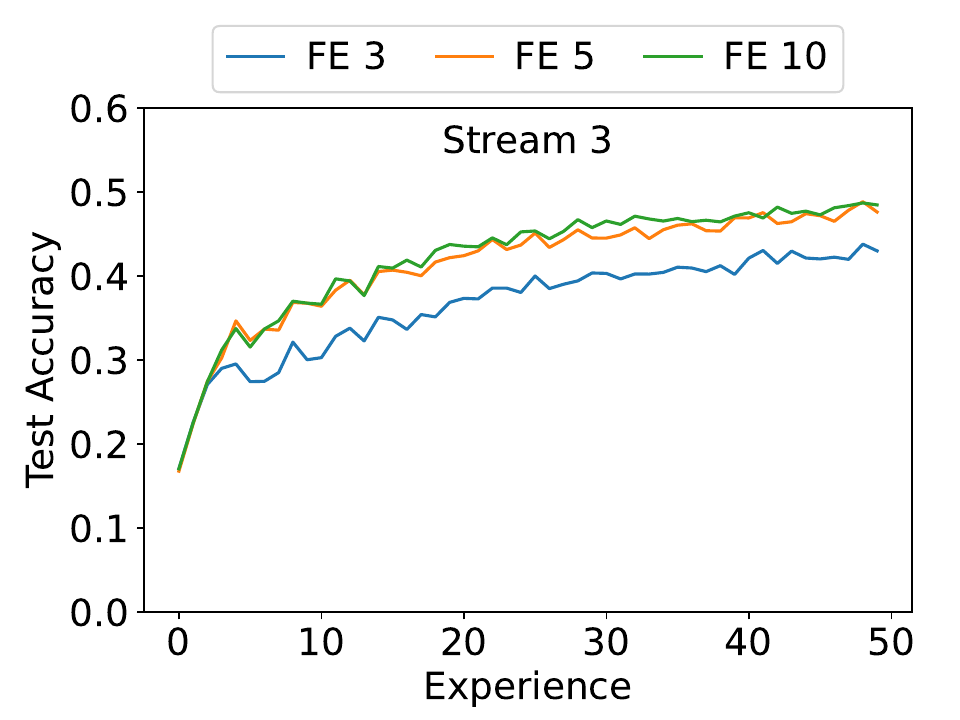}
    \caption{Impact of the maximum number of feature extractors allowed to be trained. The heuristic to decide when feature extractors are trained is not changed. Results are shown in the form of the test accuracy~(as a proportion) after training on each of the three streams from the challenge's initial phase.
    }
    \label{fig:horde-num-fes}
\end{figure}

To further study the effects of the ensemble size, the Horde strategy is evaluated with different number of maximum ensemble size. The results are presented in Figure~\ref{fig:horde-num-fes}. Stream 1 seems to be invariant to the ensemble size, probably due to the first experience having 51 classes and, therefore, providing a robust representation from the beginning. This can be exploited by heavily enforcing stability and limited number of FEs, thus becoming unnecessary to extend the ensemble after enough classes are seen. Streams 2 and 3 seem to show a benefit of extending the ensemble over time, which accommodates the incorporation of new seen classes without the need of replacing older FEs.

Another interesting direction to investigate is the different possibilities for estimating unknown class statistics for a feature extractor. Since multiple FEs exist, access might not be available to the mean and standard deviation for each class, depending on when they were learned, or if those classes have reappeared since last time. The complete class statistics $\mu_c$ and $\sigma_c$ can thus be written as a concatenation of the individual FEs where $n$ denotes the number of feature extractors.
\[
\mu_c = (\mu_{c, 1}, \, \dots \,, \mu_{c, n})
\]
\[
\sigma_c = (\sigma_{c, 1}, \, \dots \,, \sigma_{c, n})
\]

For the pseudo-feature projection step, given a class $c$ and a feature extractor $e$, estimates are needed for $\hat{\mu}_{c, e}$ and $\hat{\sigma}_{c, e}$. In cases we do not have any information about the class shape for this feature extractor, we fix the estimation of the standard deviation to $\hat{\sigma} = 1$. For the estimation of $\hat{\mu}_{c, e}$ we propose three heuristics:

\begin{enumerate}
    \item \textbf{zeros}: setting all $\hat{\mu}_{c, e}$ estimations to $0$.
    \item \textbf{random}: randomly sample $\hat{\mu}_{c, e}$ from a multivariate normal distribution $\mathcal{N}(0, 1)$.
    \item \textbf{original features}: estimate $\hat{\mu}_{c, e}$ with the original representation of the current sample without modification.
\end{enumerate}

\begin{figure}[h]
    \centering
    \includegraphics[width=0.32\textwidth, trim=10 8 20 8, clip]{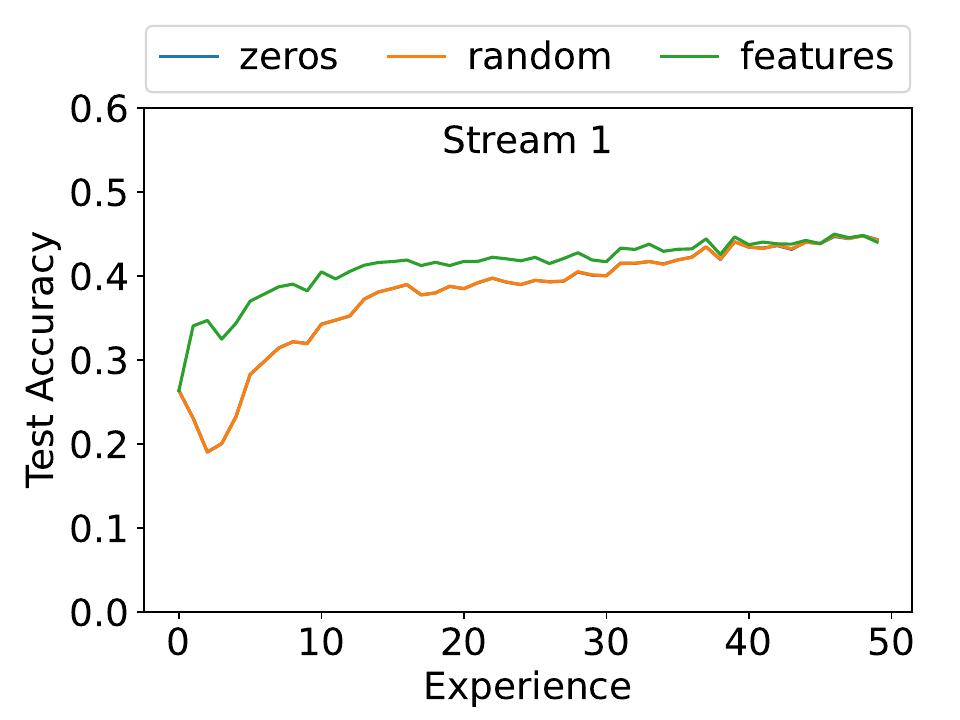}
    \includegraphics[width=0.32\textwidth, trim=10 8 20 8, clip]{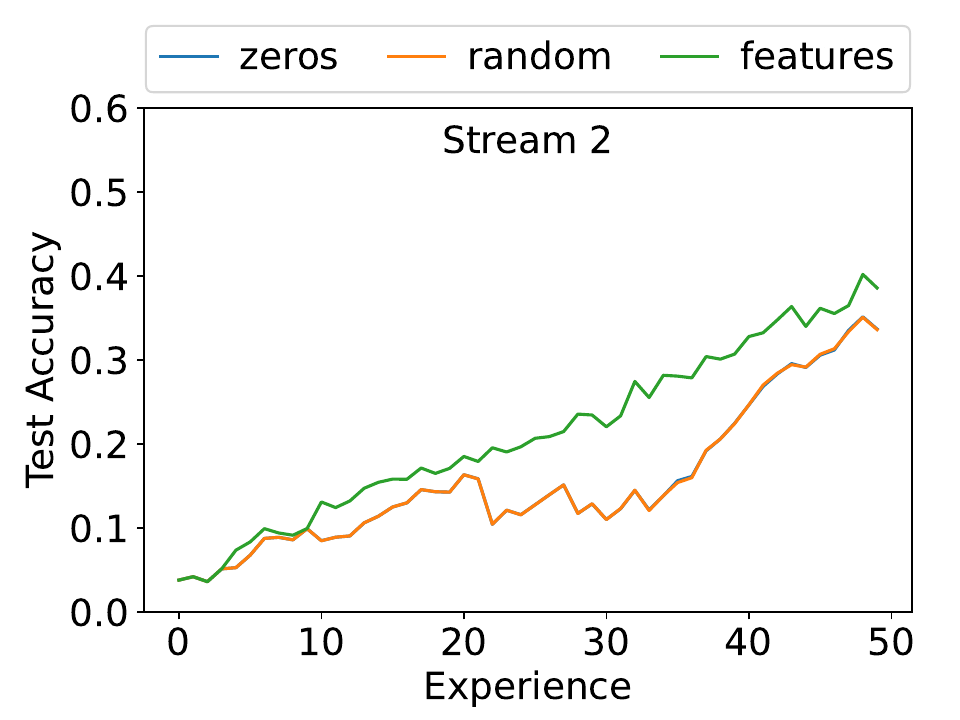}
    \includegraphics[width=0.32\textwidth, trim=10 8 20 8, clip]{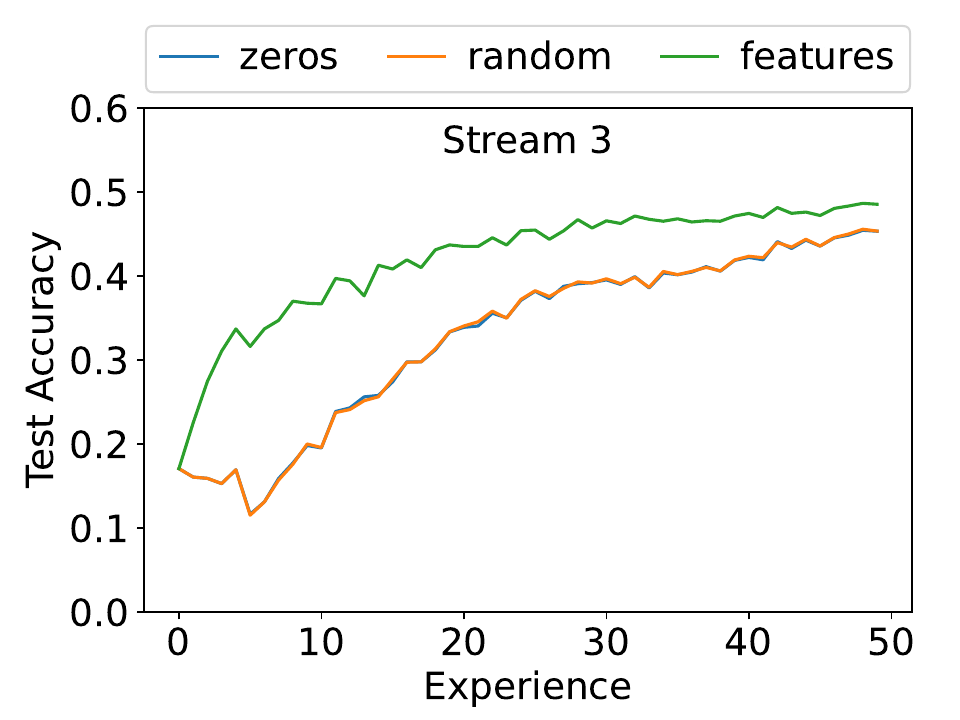}
    \caption{Impact of estimating unknown class statistics under different heuristics. Zeros and random heuristics had almost the same performance. Results are shown in the form of the test accuracy~(as a proportion) after training on each of the three data streams from the initial phase of the challenge.}
    \label{fig:horde-estimate}
\end{figure}

Results are presented in Figure~\ref{fig:horde-estimate}, showing that estimating with the original representations provides a clear advantage. Both zeros and random heuristics seem to perform similarly, and clearly underperform compared to the original features one.

\section{Appendix -- DWGRNet}\label{sec:appendix_pddbend}
\setcounter{figure}{0}
\setcounter{table}{0}

\subsection{Additional Experimental Analysis}
An ablation study was conducted to investigate the role of each module. The results are presented in Table \ref{tab:dwgrnet_final_results}. It can be seen that entropy plays the most significant role in the final accuracy. Adjusting each model's logits using the number of classes and the number of features also leads to improved accuracy. This verifies the effectiveness of each module.

\begin{table}[H]
    \centering
    \begin{tabular}{c c c c c}
    \toprule
         Module & Average Accuracy $\uparrow$ & Accuracy $S_1$ & Accuracy $S_2$ & Accuracy $S_3$  \\
         \cmidrule(lr){1-1} \cmidrule{2-5}
         
         Only mean       & $18.34$ & $20.22$ & $12.08$ & $22.71$	\\
         + Entropy	    & $41.61$ & $49.16$ & $38.14$ & $37.52$ \\
         + $N_C$	    & $42.79$ & $50.11$	& $38.62$ &	$39.64$	\\
         + feature\_norm	& $44.77$ & $52.32$	& $40.31$ &	$41.67$ \\
         
    \bottomrule
    \end{tabular}
    
    \vspace{0.2 cm}
    \caption{Result from the ablation study for the DWGRNet strategy.}
    \label{tab:dwgrnet_final_results}
\end{table}

\end{document}